\documentclass[11pt]{article}

\usepackage[preprint]{acl}

\usepackage{times}
\usepackage{latexsym}

\usepackage[T1]{fontenc}
\usepackage[utf8]{inputenc}
\usepackage{microtype}

\usepackage{graphicx}

\usepackage{booktabs}
\usepackage{tabularx}
\usepackage{multirow}
\usepackage{makecell}
\usepackage{amsmath}
\usepackage{amssymb}
\usepackage[breakable, skins]{tcolorbox}
\usepackage{caption}
\usepackage{cleveref}
\usepackage{float}

\usepackage{xcolor}
\usepackage{tcolorbox}
\usepackage{multirow}
\usepackage{booktabs}
\usepackage{makecell}
\usepackage{colortbl}
\title{Knowledge Editing for Masked Diffusion Language Models}

\author{
  Haewon Park \quad Yohan Jo$^{\dag}$
  \\
  Graduate School of Data Science, Seoul National University\\
  \texttt{\{dellaanima2, yohan.jo\}@snu.ac.kr}
}

\begin{document}
\maketitle
\renewcommand{\thefootnote}{\fnsymbol{footnote}}
\footnotetext[2]{Corresponding author.}
\renewcommand{\thefootnote}{\arabic{footnote}}   
\begin{abstract}
Knowledge editing aims to update or correct factual knowledge in a 
language model. A widely used approach, locate-then-edit, first localizes a fact within the model and then
edits the weights there. To date, such methods have been developed
exclusively for autoregressive models (ARMs). Whether they work for masked diffusion models (MDMs), which model text
bidirectionally and generate by iterative denoising rather than
next-token prediction, remains an open question. 
We address it by transferring locate-then-edit to MDMs and comparing multiple MDMs with their matched ARMs.
Our central finding has two parts. First, \emph{where} an edit should be applied
transfers between them: the same
early-to-mid-layer MLP at the last subject token is most effective for both. Second, this shared location does not
guarantee a shared outcome. Single-token edits succeed in both, but as
targets grow longer, editing degrades far more sharply in the MDMs than in the ARMs. The failure stems from \emph{how} the edited fact is
generated: producing a multi-token target passes through intermediate states in which the target is partially unmasked, for which the edit was never
optimized. Guided by this diagnosis, we introduce a simple correction
that optimizes the edit including such states, substantially restoring
multi-token performance.
Our code is available at \url{https://github.com/holi-lab/MDM-KE}.
\end{abstract}

\section{Introduction}
Large language models (LLMs) acquire extensive factual knowledge through
pretraining, but not all of it is correct: some facts are simply wrong, a
product of noise in the training data, and others grow stale over
time \citep{10.1145/3571730, lin-etal-2022-truthfulqa, jang2022towards}.
Fine-tuning can fix them, but doing so over the full model is
computationally expensive and risks overfitting to the correction while
eroding unrelated knowledge through catastrophic
forgetting \citep{bethune2025scalinglawsforgettingfinetuning, luo2025empiricalstudycatastrophicforgetting, zhu2020modifyingmemoriestransformermodels}.
Knowledge editing offers an alternative, revising targeted facts while
keeping the rest of the model's knowledge
intact \citep{yao-etal-2023-editing, wang2024knowledgeeditinglargelanguage}.
A fact is typically represented as a tuple $(s, r, o)$---a subject $s$, a
relation $r$, and an object $o$---and an edit replaces the object with a new target object $o^{*}$ for a given $(s, r)$: for example, ``Lionel
Messi $(s)$ plays for $(r)$ FC Barcelona $(o)$'' can be edited to ``Inter Miami''
$(o^{*})$. Among such methods, the locate-then-edit approach
identifies where a fact is causally localized by analyzing the model's
internals, then modifies only the weights at that
location \citep{meng2023locating, meng2023memit}. Because the edit is
grounded in an explicit analysis rather than distributed across the whole
model or delegated to an auxiliary
network \citep{decao2021editing, mitchell2022fast, mitchell2022memory}, the
edit site is interpretable.

These methods, however, were developed and validated entirely on
autoregressive models (ARMs). Masked diffusion models (MDMs)~\citep{nie2025llada,ye2025dream,zhu2025lladamoesparsemoediffusion} have
recently emerged as a competitive alternative paradigm, yet knowledge
editing for MDMs remains largely unexplored. Whether locate-then-edit,
designed for ARMs, also applies to MDMs is the question this work
takes up. The challenge runs deeper than reimplementation: the
location these methods modify is not arbitrary but derived from how
ARMs work. Using causal tracing, ROME~\citep{meng2023locating}
localized a fact's recall to the early-to-mid-layer MLP of the last
subject token, read this MLP as a form of associative memory, and
made it the edit target; MEMIT~\citep{meng2023memit} built on the same
site to scale to many edits at once. This grounding, however, comes
entirely from ARMs, which differ from MDMs even in how knowledge is
acquired: an ARM is trained to predict the next token, whereas an MDM recovers masked tokens from bidirectional context. It is
therefore not obvious that an edit site identified in ARMs carries
over to a model trained so differently.

To investigate this, we first transfer locate-then-edit to MDMs without
altering its core algorithm, applying only the adjustments needed for masked prediction (\S\ref{para:transfer}), and we compare multiple
MDMs with their matched ARMs.
We examine the question at two levels: we perform causal tracing to identify where
facts are causally localized, and we measure editing performance
across token positions and layer ranges to identify where editing is
most effective. We find that even in MDMs, where every token attends
to the full subject bidirectionally, the localization of facts
coincides with that in ARMs---the early-to-mid-layer MLP of the last
subject token. The location assumption underlying locate-then-edit is
thus not specific to autoregressive training but holds across both
paradigms.

Sharing a location, however, does not mean sharing an outcome. While
the two paradigms perform comparably for single-token targets, editing
in MDMs degrades once the target spans multiple tokens. The cause lies
not in the edit location but in a mismatch between edit optimization
and MDM generation. In an ARM, a multi-token target is generated autoregressively, with each token conditioned on its preceding target
prefix. Under the direct MDM adaptation, however, the edit is optimized using
only the fully masked output. During generation, the model instead
receives a sequence of inputs in which some target positions are
filled while others remain masked. We include such intermediate states when optimizing the edit,
substantially restoring multi-token editing in MDMs.

Our contributions are as follows:
\begin{itemize}
\item We present the \textbf{first analysis of locate-then-edit
knowledge editing in MDMs} (\S\ref{sec:transfer}), showing that its
core location assumption---that recall is mediated by the
early-to-mid-layer MLP at the last subject token---holds for both ARMs and MDMs.
\item We find that in MDMs, locate-then-edit suffers a
\textbf{multi-token breakdown} (\S\ref{sec:failure}): editing succeeds
for single-token target objects but degrades systematically as the
target grows longer. We trace this failure to a mismatch between the
fully masked state used to optimize the edit and the partially
unmasked states encountered during iterative denoising. Standard
benchmarks, dominated by single-token targets, can obscure
this breakdown, so we analyze it using a multi-token evaluation adapted
from \textsc{Kamel}~\citep{kalo2022kamel}.
\item We propose a \textbf{simple correction} (\S\ref{sec:failure-fix})
that makes the edit robust to the intermediate states traversed during
generation, substantially restoring multi-token editing performance in
MDMs.
\end{itemize}

\section{Preliminaries}
\subsection{Locate-then-Edit Knowledge Editing}
\label{sec:prelim_ke}
Knowledge editing aims to modify a fact triplet from $(s, r, o)$ to 
$(s, r, o^\ast)$. Locate-then-edit methods~\citep{meng2023locating, 
meng2023memit} achieve this by viewing the down-projection layer 
$\mathbf{W}$ of a Transformer MLP as a linear associative 
memory~\citep{anderson1972, kohonen1972} that maps a key vector 
$\mathbf{k}$ to a value vector $\mathbf{v}$ via $\mathbf{W}\mathbf{k} = 
\mathbf{v}$. The key $\mathbf{k}$, taken as the MLP up-projection 
activation at the last token of the subject $s$, encodes the entity, 
while the value $\mathbf{v}$ encodes the associated $(r, o)$. Editing a 
fact amounts to adding an update $\mathbf{\Delta}$ to $\mathbf{W}$ so 
that the same key maps to a new value $\mathbf{v}^\ast$ encoding 
$(r, o^\ast)$, i.e., $(\mathbf{W}+\mathbf{\Delta})\mathbf{k} \approx 
\mathbf{v}^\ast$. Within this general formulation, subsequent methods
differ in how the edit is constructed and applied: some constrain the
resulting parameter update to better preserve existing
knowledge~\citep{fang2025alphaedit}, while others refine the internal
representations used to construct the
edit~\citep{li2024pmet, park2026suit}. The choice of where to edit rests
on causal tracing~\citep{meng2023locating}, which identifies the
components most responsible for recalling a fact---the
early-to-mid-layer MLPs at the last subject token.

Building on this site, MEMIT~\citep{meng2023memit} edits a contiguous
range of these MLP layers and, for a batch of facts, solves for the
update $\mathbf{\Delta}$ in closed form from the keys $\mathbf{k}$ and
the corresponding target values $\mathbf{v}^\ast$ (full derivation in
Appendix~\ref{app:memit}). While the key $\mathbf{k}$ is read directly
from the subject, the target value $\mathbf{v}^\ast$ is obtained by
optimization: it is the value that, when written at the edit location,
drives the model to produce the target object $o^\ast$ given the
prompt $p$ describing $(s, r)$. Concretely, $\mathbf{v}^\ast = \mathbf{v} +
\boldsymbol{\delta}$, where the residual $\boldsymbol{\delta}$ is found
by gradient descent to maximize the probability of $o^\ast$ under the
model with the edit-site value replaced by
$\mathbf{v}+\boldsymbol{\delta}$, denoted
$G(\mathbf{v}\leftarrow\mathbf{v}+\boldsymbol{\delta})$:
\begin{equation}
\label{eq:delta_opt}
\boldsymbol{\delta} = \arg\min_{\tilde{\boldsymbol{\delta}}}\;
-\log \mathbb{P}_{G(\mathbf{v}\leftarrow
\mathbf{v}+\tilde{\boldsymbol{\delta}})}\bigl[\, o^\ast \mid p \,\bigr].
\end{equation}
The optimization details follow MEMIT~\citep{meng2023memit}
(Appendix~\ref{app:memit}). Crucially, $\boldsymbol{\delta}$ is fit for a single conditioning
state. This implicitly assumes the model encounters that same state
when it generates the target---an assumption we revisit for MDMs in
\S\ref{sec:failure}.

\subsection{Masked Diffusion Language Models}
\label{sec:prelim-mdm}
ARMs factorize the joint distribution into left-to-right conditionals,
predicting each token from its left context only through a causal
attention mask. An MDM is trained differently. Given a sequence
$x_0 = (x_0^1, \dots, x_0^L)$, a forward process independently
replaces each token with an absorbing mask state $\mathrm{M}$ with
probability $t \in [0,1]$, and the model learns a mask
predictor $p_\theta(\cdot \mid x_t)$ that recovers all masked tokens
simultaneously from the partially masked sequence $x_t$. Training
minimizes a cross-entropy loss evaluated only at masked
positions~\citep{nie2025llada}:
\begin{equation}
\begin{aligned}
  \mathcal{L}(\theta)
  = -\,\mathbb{E}_{t,\,x_0,\,x_t}\!\Bigg[
    \frac{1}{t}\sum_{i=1}^{L}
    \mathbf{1}\!\left[x_t^i = \mathrm{M}\right] \\
    \log p_\theta\!\left(x_0^i \mid x_t\right)
  \Bigg],
  \quad t \sim \mathcal{U}[0,1].
\end{aligned}
\label{eq:mdm-loss}
\end{equation}
Two properties of this formulation are central to our study. First,
the mask predictor attends to the entire input without a causal
mask, so every position is predicted from bidirectional context rather
than from a left prefix. Second, the model is trained to reconstruct
masked tokens rather than to predict the next one, so the factual
knowledge that locate-then-edit modifies is formed under a learning
objective different from that of ARMs. Our two primary MDMs realize this objective differently: LLaDA is trained from scratch, whereas Dream is initialized from the weights of an ARM (Qwen2.5) and trained
with a shifted objective that aligns its prediction with that of an
ARM~\citep{ye2025dream}.

\paragraph{Iterative denoising.}
Unlike AR decoding, which produces tokens one at a time, an MDM
generates by iterative denoising. Starting from a fully masked
output of length $L$, at each step the model predicts a token
distribution at every masked position, selects a subset of masked
positions to reveal, fills each selected position with a predicted
token, and leaves the remaining positions masked for subsequent
steps. Although different rules can be used to select this subset, a
common approach is confidence-based unmasking, which ranks masked
positions by a token-level confidence score and reveals the
highest-scoring positions first. Confidence can be measured in
different ways; by default, following prior
work~\citep{gwak-etal-2025-reward,nie2025llada}, we define it as the
probability assigned to each position's top-1 prediction. Regardless
of the selection rule, each selected position is filled with its
top-1 prediction.

Multi-token generation therefore passes through intermediate
states, with some positions filled and others masked. As we show in
\S\ref{sec:failure}, an edit optimized for a single conditioning
state may not remain effective across these states, which have no
direct counterpart in left-to-right AR generation.

\section{What Transfers: Locating and Editing Facts}
\label{sec:transfer}
\begin{figure*}[t]
    \centering
        \includegraphics[width=\textwidth]{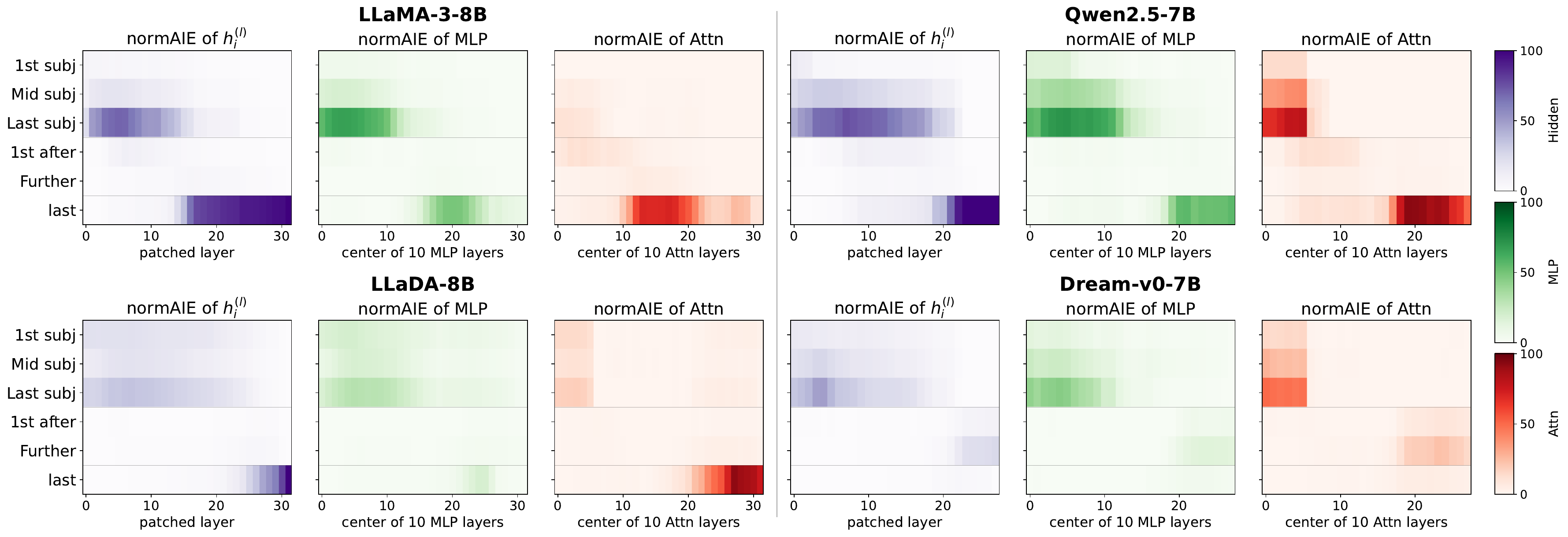}
            \caption{
            Causal tracing across ARM/MDM pairs.
            Top row: ARM baselines; bottom
            row: their MDM counterparts. `last' refers to the final input token (in MDMs, the last appended mask token).
            }
            \label{fig:causal_tracing}
                \end{figure*}

The locate-then-edit approach proceeds in two stages: causal tracing
identifies the components most responsible for recalling a fact, and
the weights at those locations are then modified. We test whether each
stage transfers to MDMs. After describing our setup
(\S\ref{sec:setup}), we first apply causal tracing to locate where facts are causally localized (\S\ref{sec:tracing}). We
then adapt locate-then-edit to MDMs and directly measure where
editing is most effective across token positions and layers
(\S\ref{sec:editing}). Across the two analyses, despite the MDMs'
bidirectional training, the same region emerges---the
early-to-mid-layer MLP at the last subject token---consistent with the
paired ARMs.

\subsection{Setup}
\label{sec:setup}

\paragraph{Models.}
Our main analyses focus on two ARM--MDM pairs matched in parameter count
and layer depth, so each comparison controls for model capacity:
LLaMA-3-8B-Instruct~\citep{grattafiori2024llama3herdmodels} (ARM) /
LLaDA-8B-Instruct~\citep{nie2025llada} (MDM), and
Qwen2.5-7B-Instruct~\citep{qwen2025qwen25technicalreport} (ARM) /
Dream-v0-Instruct-7B~\citep{ye2025dream} (MDM).
The latter is especially controlled, as Dream is initialized from Qwen2.5.

To further assess the generality of the patterns observed in our main
analyses, we conduct additional experiments on three ARM--MDM pairs:
LLaMA-3-8B-Instruct /
LLaDA-1.5-8B~\citep{zhu-etal-2026-llada},
LLaMA-2-7B~\citep{touvron2023llama2openfoundation} /
DiffuLLaMA-7B~\citep{gong2025scalingdiffusionlanguagemodels}, and
GPT-2-medium~\citep{radford2019language} /
DiffuGPT-m (355M)~\citep{gong2025scalingdiffusionlanguagemodels}.
Detailed experimental results are reported in
Appendix~\ref{app:generality_models}.

\paragraph{Dataset and metrics.}
We evaluate on \textsc{CounterFact}~\citep{meng2023locating}, the standard
knowledge-editing benchmark, using its three criteria
\citep{meng2023locating, meng2023memit}: \emph{efficacy} (the rewrite prompt \((s, r)\) elicits \(o^*\)), \emph{generalization} (paraphrase prompts also elicit \(o^*\)), and \emph{specificity} (a neighborhood prompt, with a distinct
subject $s'$ sharing $r$ and $o$, still yields the original object $o$).
Because specificity is generation-based, its absolute value reflects
not only preservation but also the model's pre-edit ability to recall
the neighboring facts. It should therefore be interpreted relative
to the pre-edit score rather than as an absolute measure of
preservation (Appendix~\ref{app:locality}).
Following MEMIT's batched setting, we edit 500 facts in a single batch.
We assess all three from the model's generation; details are deferred
to Appendix~\ref{app:eval}.

\subsection{Locating Facts via Causal Tracing}
\label{sec:tracing}

\paragraph{Causal tracing.}
To localize where subject information is processed en route to the
prediction, we follow the causal tracing procedure of
\citet{meng2023locating}: we corrupt the subject tokens and measure
how much restoring an individual mediator to its clean value recovers
the correct prediction. For each fact, we compare three runs.
The \emph{clean run} passes the unperturbed input $x$ through the model, records the hidden state $h_i^{(l)}$ at every token position $i$ and layer $l$, and produces the correct answer $o$.
The \emph{corrupted run} adds Gaussian noise
$\epsilon\sim\mathcal{N}(0,\nu^2)$ to the embeddings of all tokens
spanning the subject entity, with $\nu=3\sigma$, where $\sigma$ is the
standard deviation of subject-token embeddings, estimated once per
model. This corruption removes subject information and typically
disrupts the correct prediction.
The \emph{corrupted-with-restoration run} uses the corrupted input and selects a token position $\hat{i}$ and layer $\hat{l}$ for restoration. Writing each layer's update as
$h_i^{(l)}=h_i^{(l-1)}+a_i^{(l)}+m_i^{(l)}$, where $a_i^{(l)}$ and $m_i^{(l)}$ denote the attention and MLP contributions, respectively, we restore $h_{\hat{i}}^{(\hat{l})}$, $a_{\hat{i}}^{(\hat{l})}$, or $m_{\hat{i}}^{(\hat{l})}$ to its value from the clean run.
Because these per-layer contributions are
small, restoring a single layer has little effect, so we restore a
contiguous window of layers around $\hat{l}$ at token $\hat{i}$ when
tracing sub-modules, following \citet{meng2023locating}.
The \emph{indirect effect} (IE) is the increase in $P(o)$ upon
restoration relative to the corrupted run, and its average over facts
yields the \emph{average indirect effect} (AIE). Because the corruption-induced gap varies widely across models, we
report normalized AIE, computed for each fact as
\[
100\times
\frac{P_{\text{restored}}(o)-P_{\text{corrupted}}(o)}
     {P_{\text{clean}}(o)-P_{\text{corrupted}}(o)}
\]
and then averaged over facts. This measures the fraction of the gap
recovered rather than the absolute gain, making results comparable
across models (Appendix~\ref{app:tracing}). A value of $0$
indicates no recovery beyond the corrupted run, whereas $100\%$
indicates full recovery of the clean prediction.

We apply this procedure to all four models to identify where across
layers and sub-modules (MLP or attention) the effect concentrates.
Because MDMs predict at mask rather than next-token positions, we
append a mask block and read $P(o)$ at its first position, where the
first token of $o$ is predicted. Noise levels, restoration-window and
mask-block sizes, the MDM adaptation, and other tracing details are
provided in Appendix~\ref{app:tracing}.

\paragraph{Results.}
Figure~\ref{fig:causal_tracing} reveals two causal sites across all
four models: an early site at the last subject token in the
early-to-mid layers, carried mainly by the MLP, and a late site
at the answer position in the upper layers, dominated by attention.
The late site is expected because the prediction is read out there;
our focus is the early site, which \citet{meng2023locating} identify
as the site of factual recall. Despite bidirectional training, LLaDA
and Dream exhibit the same early MLP site as the ARMs.

The MDM peaks, however, are weaker and more diffuse. Because
localization tends to be sharper for high-confidence facts, this
difference could simply reflect differences in the models'
$P_{\text{clean}}(o)$ distributions. We therefore compare each
AR--MDM pair within matched $P_{\text{clean}}(o)$ bins. The MDM peak
remains weaker in every bin (Appendix~\ref{app:tracing}), indicating
that the location of recall is shared while recall in MDMs is more
distributed.
Qwen and Dream additionally show a secondary early-layer attention
contribution, consistent with Dream's initialization from Qwen; the
primary early-MLP site, however, is shared across all four models.
Two complementary analyses support this localization:
activation$\times$gradient attribution identifies the same
token--layer site, while MLP knockout confirms the early-to-mid-layer
concentration at the last subject token. Results are provided in
Appendix~\ref{app:additional_localization}.

\subsection{Editing Facts}
\label{sec:editing}

Causal tracing (\S\ref{sec:tracing}) identifies where a fact is
causally localized, but this need not be the most effective site to
edit~\citep{hase2023}. We therefore measure editing performance across
token positions and layers and test whether the resulting edits remain
robust across generation settings reflecting MDM decoding.

\paragraph{Transferring locate-then-edit to MDMs.}
\label{para:transfer}
For the experiments in this section, we use MEMIT as our
locate-then-edit method. While keeping its editing algorithm, we make only the minimal adjustments needed for the MDM's masked
prediction. The optimization objective for the target value remains
the same (Eq.~\ref{eq:delta_opt}); what changes is the conditioning
input under which $o^\ast$ is scored. An ARM appends the target
sequence to the rewriting prompt and, under causal attention, scores
each target token conditioned on the preceding target tokens. An MDM
instead predicts masked positions. We place one mask token at the
answer slot for each token in the edit target $o^\ast$ and optimize
the target value so that the model predicts all target tokens from
this fully masked input. For instance, to edit Messi's team to
``Inter Miami'', we place two mask tokens at the answer slot and score
``Inter'' and ``Miami'' at their respective positions from the same
fully masked state. Because an MDM uses bidirectional attention, the appended masks can influence the hidden state at the last subject token even though they appear later in the input. We therefore use this same fully masked input when constructing both components of the edit: we optimize the target value $v^*$ under this input and extract the key $k$ at its last subject token. This input matches the initial state of MDM generation. We refer to this MDM adaptation as MEMIT$^\dagger$; the full procedure is provided in Appendix~\ref{app:editing}.

\begin{figure}[t]
\centering
\includegraphics[width=\columnwidth]{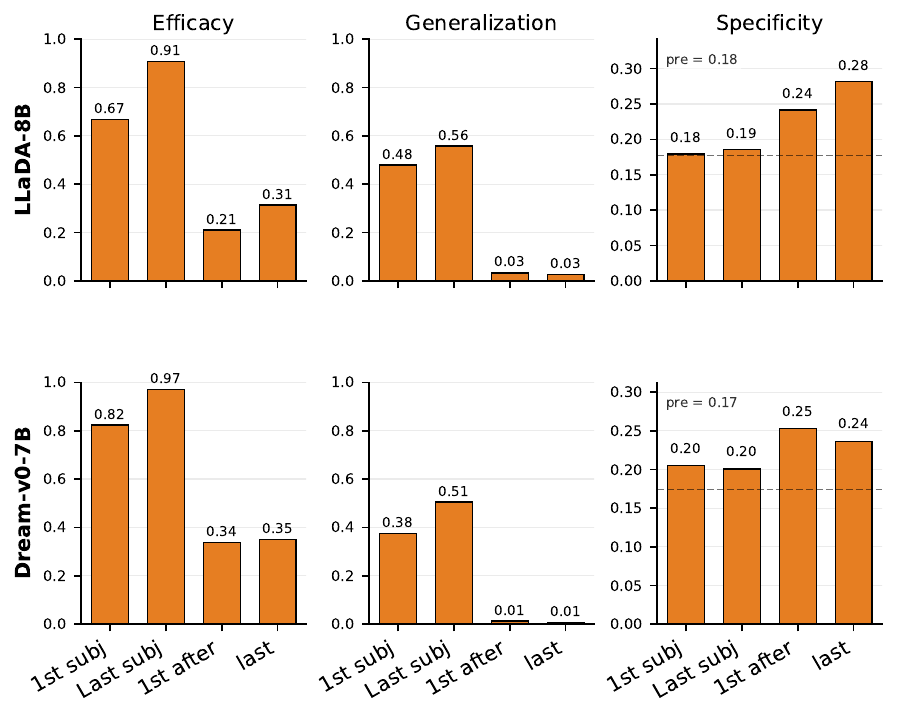}
\caption{
Token position ablation for editing.
}
\label{fig:token_sweep}
\end{figure}

\begin{figure}[t]
    \centering
        \includegraphics[width=\columnwidth]{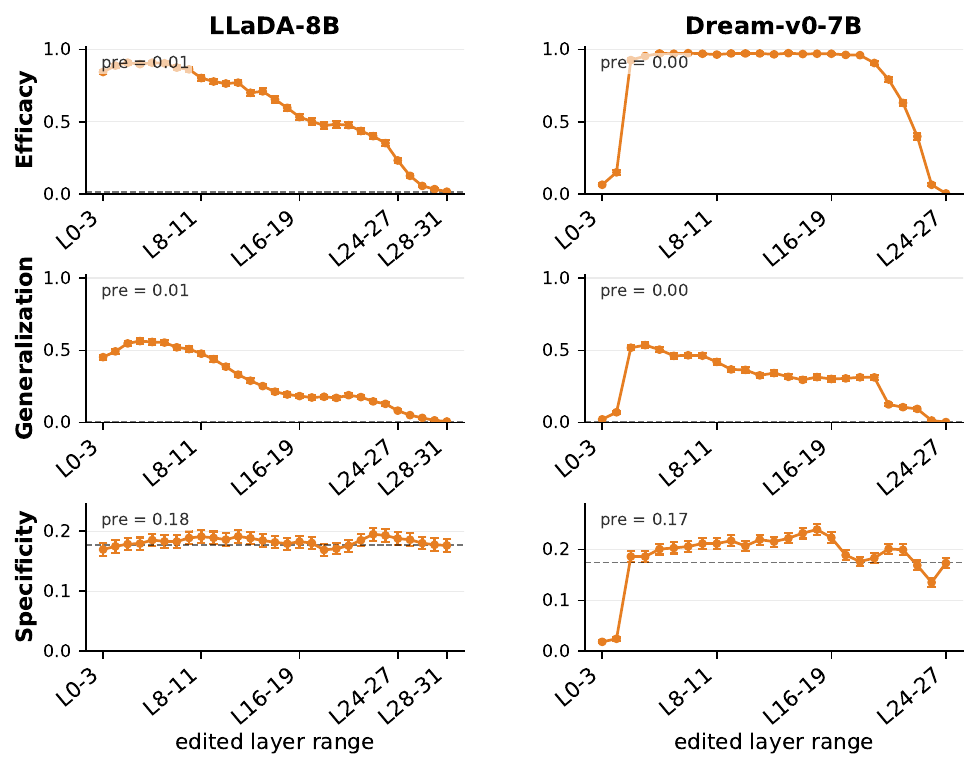}
            \caption{Layer ablation for editing.\label{fig:layer_sweep}}
            \end{figure}

\paragraph{Token position and layer.}
\label{sec:editing-token}
\label{sec:editing-layer}
We vary both the token position and layer range at which the edit is
applied (Figures~\ref{fig:token_sweep} and~\ref{fig:layer_sweep}).
The tested positions are the first and last subject tokens, the first
post-subject token, and the first answer position---the last prompt
token for an ARM and the first appended mask for an MDM. In both MDMs,
editing peaks at the last subject token, remains strong at the first,
and drops sharply after the subject. This pattern matches both the ARM
recall site reported by \citet{meng2023locating} and our causal-tracing
results, despite the MDMs' bidirectional attention. Across layers,
editing succeeds broadly, with the strongest efficacy and
generalization in the early-to-mid range. This range overlaps but does
not exactly coincide with the causal-tracing site, consistent with
prior findings~\citep{hase2023}; we therefore select each model's edit
layers by direct measurement. ARM results and layer-selection details
are provided in Appendices~\ref{app:arm_editing}
and~\ref{app:layer_selection}, respectively, with results for
additional ARM--MDM pairs in Appendix~\ref{app:generality_models}.

\paragraph{Robustness and comparison with fine-tuning.}
\label{sec:editing-robust}
Finally, we test whether the selected edits survive changes in MDM
decoding, which varies with both generation length and denoising
steps. Although each edit is optimized only at the target length
(\S\ref{para:transfer}), it remains effective when both are varied
over $\{8,16,32\}$ at evaluation: efficacy ranges from $0.82$ to
$0.86$ for LLaDA and from $0.93$ to $0.95$ for Dream, with no
consistent degradation in generalization or specificity
(Appendix~\ref{app:generation_robustness}). Thus, the edits transfer
to decoding configurations on which they were not optimized.
Separately, we also compare MEMIT$^\dagger$ with constrained
fine-tuning on LLaDA and find that MEMIT$^\dagger$ generalizes
substantially better to paraphrased prompts (Appendix~\ref{app:ft}).

\section{What Breaks: Editing Multi-Token Facts}
\label{sec:failure}

The previous section showed that factual recall in MDMs follows the
same pattern as in their paired ARMs: it is concentrated in the
early-to-mid-layer MLPs at the last subject token, where editing is
also effective. However, these editing results were obtained on
\textsc{CounterFact}, where most target objects are tokenized as a
single token.

For a single-token target, revealing its only token completes
generation, so the fully masked state used for edit optimization is
also the only state in which the model predicts the target. For a
multi-token target, however, iterative denoising passes through
partially unmasked states that are not included in edit optimization,
raising a potential mismatch between optimization and generation.
ARMs do not face this particular mismatch: their edit objective
conditions each target token on the preceding target tokens, matching
the left-to-right conditioning used during generation
(\S\ref{para:transfer}).

In this section, we first show that editing performance degrades as
target length increases (\S\ref{sec:failure-degrade}), diagnose the
source of this degradation (\S\ref{sec:failure-diagnose}), and
introduce an intermediate-state correction that recovers much of the
lost performance (\S\ref{sec:failure-fix}). We then extend the analysis to two other locate-then-edit methods, observing the same breakdown and consistent gains from the correction
(\S\ref{sec:method-generality}).

\subsection{Editing Degrades with Target Length}
\label{sec:failure-degrade}

\begin{table}[t]
\centering
\small
\setlength{\tabcolsep}{5pt}
\begin{tabular}{@{}lrrrr@{}}
\toprule
& \multicolumn{2}{c}{\textsc{CounterFact}}
& \multicolumn{2}{c}{\textsc{Kamel}}
\tabularnewline
\cmidrule(lr){2-3}\cmidrule(lr){4-5}
Target length & LLaDA & LLaMA & LLaDA & LLaMA
\tabularnewline
\midrule
1 token
& 20{,}442 & 20{,}534 & 8{,}815 & 8{,}686
\tabularnewline
2 tokens
& 286 & 343 & 12{,}055 & 11{,}912
\tabularnewline
3 tokens
& 149 & 0 & 11{,}187 & 11{,}072
\tabularnewline
4+ tokens
& 0 & 0 & 21{,}230 & 21{,}617
\tabularnewline
\midrule
Total
& 20{,}877 & 20{,}877 & 53{,}287 & 53{,}287
\tabularnewline
\bottomrule
\end{tabular}
\caption{Distribution of target token lengths under each model's
tokenizer. Results for Dream and Qwen are reported in
Appendix~\ref{app:kamel}.}
\label{tab:length_dist}
\end{table}

Table~\ref{tab:length_dist} shows that most target objects in
\textsc{CounterFact} are tokenized as a single token. Although
multi-token targets account for only a small fraction of the dataset,
their failure rate is substantially higher than that of single-token
targets (Appendix~\ref{app:length_failure}). This provides an initial
indication that editing becomes harder as target length increases, but
\textsc{CounterFact} contains too few multi-token examples for a
systematic comparison across lengths.

To study target length systematically, we use
\textsc{Kamel}~\citep{kalo2022kamel}, a Wikidata-based probing dataset
whose objects span a wide range of token lengths. We adapt its facts to our editing setup by recasting them into the \textsc{CounterFact}
schema and selecting another object from the same relation as the
counterfactual target. For each $N\in\{1,2,3,4\}$, we sample 200 facts
whose target objects have the same token length under both tokenizers
within each ARM--MDM pair; full construction and filtering details are
provided in Appendix~\ref{app:kamel}. Unlike the robustness analysis
in \S\ref{sec:editing-robust}, which varies the output budget for a
fixed target, this analysis varies the length of the target itself.

\begin{table}[t]
\centering
\small
\setlength{\tabcolsep}{8pt}
\begin{tabular}{@{}ccccc@{}}
\toprule
$N$ & LLaMA & LLaDA & Qwen & Dream
\tabularnewline
\midrule
1 & 0.98 & 0.92 & 1.00 & 0.97
\tabularnewline
2 & 0.95 & 0.60 & 0.99 & 0.88
\tabularnewline
3 & 0.92 & 0.33 & 0.97 & 0.47
\tabularnewline
4 & 0.89 & 0.27 & 0.95 & 0.34
\tabularnewline
\bottomrule
\end{tabular}
\caption{Editing efficacy on \textsc{Kamel} by target token length
$N$ ($n{=}200$ facts per model pair at each $N$).}
\label{tab:kamel_length}
\end{table}

All four models achieve high editing efficacy on single-token targets
(Table~\ref{tab:kamel_length}). As target length increases, however,
LLaMA declines only from $0.98$ to $0.89$ and Qwen remains at or above
$0.95$, whereas LLaDA falls from $0.92$ to $0.27$ and Dream from
$0.97$ to $0.34$. Within each ARM--MDM pair, both models are evaluated
on the same facts at each length, so the widening gap cannot be
attributed to differences in fact selection
(Appendix~\ref{app:kamel}).

\subsection{The Edit Is Not Optimized for Intermediate States}
\label{sec:failure-diagnose}

To determine why editing efficacy degrades with target length
(\S\ref{sec:failure-degrade}), we examine the conditioning states
encountered during generation. MEMIT$^\dagger$ optimizes the target
value only in the fully masked state, whereas multi-token generation
passes through partially unmasked states. We therefore test whether
an edit that succeeds in the fully masked state remains effective as
target tokens are progressively revealed.

We first evaluate this mismatch under controlled changes to the
conditioning state. For a target $(t_1,\ldots,t_N)$, let $S_j$ denote
the state in which the first $j$ gold target tokens are revealed and
the remaining target positions are masked. Each $S_j$ is evaluated
independently with a separate forward pass. We restrict the analysis
to cases in which the top-1 prediction at every target position in
$S_0$ matches the corresponding gold token. Starting from these cases,
we measure whether revealing one or two gold target tokens causes any
top-1 prediction in the remaining masked suffix to become incorrect
at $S_1$ or $S_2$. Because the revealed positions contain gold rather
than model-generated tokens, this test isolates the effect of the
conditioning-state change from error propagation during generation.
After one gold token is revealed at $S_1$, at least one token in the
remaining suffix becomes incorrect in 33.1--53.4\% of cases that
succeed at $S_0$, with the rate exceeding half for $N{=}4$
(Table~\ref{tab:failure-analyses}, left). At $S_2$, the corresponding
rates fall to about 19\% for both $N{=}3$ and $N{=}4$. Even so,
revealing a single correct target token can disrupt an edit that
succeeds in the fully masked state.

We next ask whether a corresponding pattern appears in actual
generation outputs from the target-length evaluation
(Table~\ref{tab:kamel_length}). Among failed generations, we define a
partial failure as one that contains at least one target token in its
correct position, and a complete miss as one that contains none.
Partial failures constitute the majority at every multi-token target
length (59.9--74.1\%; Table~\ref{tab:failure-analyses}, right). This
generation-level pattern is consistent with the controlled result:
an edit can predict all target tokens correctly in the fully masked
state yet lose only part of the target as generation proceeds through
intermediate states unseen during optimization.

\begin{table}[t]
\centering
\small
\setlength{\tabcolsep}{5pt}
\begin{tabular}{@{}c rr @{\hspace{10pt}} rr@{}}
\toprule
& \multicolumn{2}{c}{Gold-prefix control}
& \multicolumn{2}{c}{Generation failures} \\
\cmidrule(lr){2-3}\cmidrule(lr){4-5}
$N$ & $S_1$ & $S_2$ & Partial & Miss \\
\midrule
2 & 33.1\% & --     & 74.1\% & 25.9\% \\
3 & 35.1\% & 19.5\% & 63.0\% & 37.0\% \\
4 & 53.4\% & 19.0\% & 59.9\% & 40.1\% \\
\bottomrule
\end{tabular}
\caption{Controlled and generation-level analyses of multi-token
editing failures on \textsc{Kamel} with LLaDA
($n{=}200$ per $N$). Left: conditional suffix failure rates at $S_j$
among cases successful at $S_0$. Right: partial failures versus
complete misses among failed generations.}
\label{tab:failure-analyses}
\end{table}

\subsection{A Correction for Intermediate States}
\label{sec:failure-fix}

\paragraph{Method and performance recovery.}
\S\ref{sec:failure-diagnose} showed that a residual
$\boldsymbol{\delta}$ optimized only in the fully masked state can
lose its effectiveness once target positions are revealed. We
therefore include partially unmasked target states alongside the fully
masked state when optimizing $\boldsymbol{\delta}$. This correction
changes the conditioning states used for target-value
optimization; the weight-update rule and edited layers remain
unchanged.

Concretely, let the $N$-token target be
$(t_1,\ldots,t_N)$. At the first optimization step, no target positions
are revealed. Thereafter, we uniformly sample the number of revealed
positions from $\{0,\ldots,N-1\}$ and then uniformly sample a subset
$R\subseteq\{1,\ldots,N\}$ of that size. For each $i\in R$, we reveal
the gold token $t_i$, while leaving all positions outside $R$ masked.
We then optimize $\boldsymbol{\delta}$ to predict the corresponding gold token at
every remaining masked position.
As a result, $\boldsymbol{\delta}$ is optimized over the fully masked
state and a range of partially unmasked states that vary in both the
number and locations of revealed tokens. Because each state is sampled
independently, the correction does not assume a fixed reveal order or
depend on a particular decoding trajectory.

As shown in Table~\ref{tab:correction}, the correction substantially
mitigates the multi-token performance drop. Editing efficacy and
generalization improve across target lengths on both LLaDA and
Dream. Most notably, LLaDA's efficacy at $N{=}4$ increases from 0.27
to 0.73, nearly tripling, while Dream's generalization at $N{=}2$ increases from 0.39 to 0.80. The same pattern extends beyond
the two ARM--MDM pairs in our main analysis: across three additional
pairs, performance again degrades with target length without the
correction and recovers when the correction is applied
(Appendix~\ref{app:generality_models}). 
Moreover, broader evaluations show that the correction remains broadly
comparable to MEMIT$^\dagger$ in general capability, generation
quality, and specificity across edit scales, and further examine its
effects on related knowledge (Appendix~\ref{app:locality}).

\begin{table}[t]
\centering
\small
\setlength{\tabcolsep}{4pt}
\begin{tabular}{@{}ccrrrr@{}}
\toprule
& & \multicolumn{2}{c}{LLaDA}
    & \multicolumn{2}{c}{Dream} \\
\cmidrule(lr){3-4}\cmidrule(lr){5-6}
$N$ & Metric
& MEMIT$^\dagger$ & +Corr.
& MEMIT$^\dagger$ & +Corr. \\
\midrule
\multirow{2}{*}{2}
& Eff. & 0.60 & 0.87 & 0.88 & 0.99 \\
& Gen. & 0.36 & 0.52 & 0.39 & 0.80 \\
\midrule
\multirow{2}{*}{3}
& Eff. & 0.33 & 0.76 & 0.47 & 0.90 \\
& Gen. & 0.20 & 0.36 & 0.17 & 0.68 \\
\midrule
\multirow{2}{*}{4}
& Eff. & 0.27 & 0.73 & 0.34 & 0.80 \\
& Gen. & 0.14 & 0.36 & 0.20 & 0.60 \\
\bottomrule
\end{tabular}
\caption{Editing performance on \textsc{Kamel} for multi-token targets
($n{=}200$ per $N$). MEMIT$^\dagger$ denotes our MDM adaptation of
MEMIT (\S\ref{para:transfer}); \emph{$+$ correction} fits the edit
across partially unmasked states rather than the fully masked one
alone.}
\label{tab:correction}
\end{table}

\paragraph{Ablations on intermediate-state sampling.}
We ablate two choices in sampling intermediate states: how many target
tokens are revealed and which positions are revealed, with full
results reported in Appendix~\ref{app:augmentation}. Biasing the number of revealed positions toward smaller, larger, or intermediate values provides no consistent advantage over uniform sampling. For position selection, uniformly
resampling the revealed positions outperforms both a fixed
left-to-right policy and a confidence-based policy that reveals the
positions assigned the highest top-1 probabilities by the unedited
model. For example, at $N{=}3$, random resampling improves efficacy
over the next-best reveal policy by $14.5$ percentage points on LLaDA
and $5.0$ percentage points on Dream. Its generalization scores are
also $2.0$ and $8.5$ percentage points higher than the next-best
scores, respectively. These results suggest that the correction
benefits from exposure to diverse position combinations rather than
fitting $\boldsymbol{\delta}$ to a particular reveal pattern or
decoding trajectory.

\paragraph{Robustness across decoding strategies.}
Our main evaluation selects positions by top-1 probability and reveals
one position per step. We test robustness to both components of this
decoding strategy: which position is selected and how many positions
are revealed at each step. In all settings, selected positions are
filled with their top-1 predictions.
First, we fix the reveal count to one position per step and compare
five position-selection rules: top-1 probability, top-1--top-2 margin,
minimum entropy, random selection, and left-to-right order
(Table~\ref{tab:position-policy}). The correction consistently
outperforms MEMIT$^\dagger$ under all five rules. For $N{=}2$,
efficacy ranges from $0.54$ to $0.60$ with MEMIT$^\dagger$ and from
$0.86$ to $0.89$ with the correction. For $N{=}4$, the corresponding
ranges are $0.25$--$0.28$ and $0.70$--$0.75$.

Second, we fix top-1 probability as the position-selection rule and
vary the number of positions revealed per step
(Table~\ref{tab:reveal-counts}). Each tuple specifies the reveal counts
at successive steps; for example, $(2,1)$ for $N{=}3$ reveals two
positions in the first step and the remaining position in the second.
The correction again outperforms MEMIT$^\dagger$ under every tested
schedule. For $N{=}3$, efficacy is $0.32$--$0.40$ with
MEMIT$^\dagger$ and $0.69$--$0.78$ with the correction across the
tested schedules; for $N{=}4$, the corresponding ranges are
$0.22$--$0.33$ and $0.72$--$0.74$. These results show that the
correction remains effective regardless of which positions are
selected or how many are revealed at each decoding step.


\begin{table}[t]
\centering
\small
\setlength{\tabcolsep}{3.5pt}
\begin{tabular}{@{}lrrrr@{}}
\toprule
& \multicolumn{2}{c}{$N{=}2$}
& \multicolumn{2}{c}{$N{=}4$} \\
\cmidrule(lr){2-3}\cmidrule(lr){4-5}
Rule & MEMIT$^\dagger$ & +Corr.
     & MEMIT$^\dagger$ & +Corr. \\
\midrule
Top-1 prob.   & 0.60 & 0.87 & 0.27 & 0.73 \\
Margin        & 0.60 & 0.87 & 0.28 & 0.72 \\
Min. entropy  & 0.60 & 0.87 & 0.26 & 0.72 \\
Random        & 0.60 & 0.89 & 0.28 & 0.70 \\
Left-to-right & 0.54 & 0.86 & 0.25 & 0.75 \\
\bottomrule
\end{tabular}
\caption{LLaDA editing efficacy across position-selection rules.}
\label{tab:position-policy}
\end{table}

\begin{table}[t]
\centering
\small
\setlength{\tabcolsep}{1.25pt}
\begin{tabular}{@{}clrr@{\hspace{5pt}}clrr@{}}
\toprule
$N$ & Counts & MEMIT$^\dagger$ & +Corr.
& $N$ & Counts & MEMIT$^\dagger$ & +Corr. \\
\midrule
2 & $(2)$   & 0.74 & 0.92
& 4 & $(2,2)$ & 0.28 & 0.74 \\
3 & $(2,1)$ & 0.34 & 0.76
&   & $(3,1)$ & 0.33 & 0.72 \\
  & $(1,2)$ & 0.32 & 0.69
&   & $(1,3)$ & 0.22 & 0.72 \\
  & $(3)$   & 0.40 & 0.78
&   & $(4)$ & 0.31 & 0.72 \\
\bottomrule
\end{tabular}
\caption{LLaDA editing efficacy with different numbers of positions
revealed per decoding step.
Each tuple lists the reveal counts at successive steps.}
\label{tab:reveal-counts}
\end{table}

\subsection{Generality Across Editing Methods}
\label{sec:method-generality}

Our intermediate-state correction modifies target-value optimization
in locate-then-edit methods. In MEMIT$^\dagger$, it optimizes the
target value on partially unmasked inputs in addition to the fully
masked input, while leaving the final weight update unchanged. As
shown in \S\ref{sec:failure-fix}, this improves MEMIT$^\dagger$ on
multi-token targets.

To determine whether this benefit is specific to MEMIT, we evaluate
AlphaEdit~\citep{fang2025alphaedit} and
PMET~\citep{li2024pmet}, which extend MEMIT in different ways.
AlphaEdit retains MEMIT's target-value optimization but constrains the
final weight update to the null space of preserved knowledge, whereas
PMET jointly optimizes attention and MLP residuals and uses only the
optimized MLP residual to update the weights. We apply the masked-input
adaptation from \S\ref{para:transfer} to both methods and, for
multi-token targets, include the same partially unmasked inputs while
leaving other components unchanged.

\begin{table}[t]
\centering
\small
\setlength{\tabcolsep}{3.5pt}
\begin{tabular}{@{}llrrrr@{}}
\toprule
Model & Method & $N{=}1$ & $N{=}2$ & $N{=}3$ & $N{=}4$
\tabularnewline
\midrule
LLaDA
& AlphaEdit$^\dagger$ & 0.93 & 0.61 & 0.35 & 0.34
\tabularnewline
& \quad +Corr. & -- & 0.95 & 0.91 & 0.91
\tabularnewline
& PMET$^\dagger$ & 0.95 & 0.73 & 0.58 & 0.57
\tabularnewline
& \quad +Corr. & -- & 0.75 & 0.73 & 0.67
\tabularnewline
\midrule
Dream
& AlphaEdit$^\dagger$ & 1.00 & 0.99 & 0.56 & 0.46
\tabularnewline
& \quad +Corr. & -- & 0.98 & 0.97 & 0.84
\tabularnewline
& PMET$^\dagger$ & 0.96 & 0.81 & 0.67 & 0.43
\tabularnewline
& \quad +Corr. & -- & 0.90 & 0.78 & 0.61
\tabularnewline
\bottomrule
\end{tabular}
\caption{Editing efficacy of AlphaEdit and PMET across
\textsc{Kamel} target lengths ($n{=}200$ per $N$).
AlphaEdit$^\dagger$ and PMET$^\dagger$ denote our MDM adaptations;
+Corr. is applied to multi-token targets.}
\label{tab:method-generality}
\end{table}

Without the correction, efficacy declines with target length for both
methods on both models (Table~\ref{tab:method-generality}). At
$N{=}4$, the correction raises AlphaEdit from $0.34$ to $0.91$ on
LLaDA and from $0.46$ to $0.84$ on Dream, and PMET from $0.57$ to
$0.67$ and from $0.43$ to $0.61$. This recovery, though larger for
AlphaEdit, shows that the mismatch is not specific to MEMIT's original
formulation.
The difference in gains is consistent with how each method converts
the optimized target into a weight update. AlphaEdit changes only the
final solve, passing the corrected target value into the
null-space-constrained update. PMET writes only the optimized MLP
residual to model weights; the jointly optimized attention residual is
not directly retained, which may explain its smaller gains.

\section{Related Work}
\label{sec:related}

\paragraph{Knowledge Editing.}
Knowledge editing methods fall into three paradigms.
\emph{Memory-based approaches} store edits externally and retrieve
them at inference, leaving the model unchanged
\citep{mitchell2022memory, hartvigsen2023aginggracelifelongmodel}.
\emph{Meta-learning approaches} train auxiliary networks to predict
weight updates \citep{decao2021editing, mitchell2022fast}.
\emph{Locate-then-edit approaches} identify and directly modify the
parameters associated with a fact
\citep{meng2023locating, meng2023memit}. Grounded in causal
localization, this family provides interpretable edit sites and has
scaled to simultaneous edits with reduced side effects
\citep{fang2025alphaedit, li2024pmet, park2026suit,
gupta2024unifiedframeworkmodelediting, park-etal-2025-context}.
Its scalability and causally derived edit locations make it a natural
starting point for testing whether locate-then-edit transfers from ARMs
to MDMs.

\paragraph{ARMs vs.\ MDMs.}
Prior comparisons report MDM advantages in knowledge acquisition
\citep{pan2025diffusion} and differences in long-context behavior
\citep{liu2025longllada}. Closer to our setting,
\citet{wang2025masksdistracting} show that appended mask tokens impair
context comprehension and propose a mask-agnostic loss, while
\citet{wang2026dlmscope} interpret MDM representations through sparse
features. These studies focus on representations or inference-time
behavior rather than factual localization or editing. We instead
transfer locate-then-edit from ARMs to MDMs without modifying training
and compare factual localization and editing across the two paradigms.

\section{Conclusion}
\label{sec:conclusion}
We transferred locate-then-edit methods from ARMs to MDMs. The two
agree on \emph{where} editing applies: in both, facts concentrate at
the early-to-mid-layer MLP of the last subject token, and editing there
matches ARM performance on standard benchmarks. They diverge on
\emph{how} the edited fact is generated. These almost entirely
single-token benchmarks obscure this divergence; MDM editing degrades
as targets grow because denoising visits intermediate states absent
from optimization. Including these states restores most lost
performance, confirming the cause, though a gap at the longest targets
remains for future work. Thus, porting an editing method across
architectures requires matching the edit site and the generation
process.

\section*{Limitations}
\label{sec:limitations}

We evaluate several representative locate-then-edit methods, but do not cover all variants in this family. While our analysis spans multiple ARM--MDM pairs and model scales, it remains limited to dense architectures; extending the analysis to larger MDMs based on sparse mixture-of-experts architectures remains future work. Finally, although we identify the mismatch between single-state optimization and the intermediate states encountered during generation as an important cause of multi-token failure, and our correction recovers most of the lost performance by sampling partial-mask states with different numbers and subsets of revealed positions, a gap remains at the longest targets. Developing a method that more fully accounts for intermediate-state
variation without tying the edit to a particular unmasking strategy
remains future work.

\section*{Ethics Statement}
Knowledge editing modifies the facts a model produces, and the same
techniques that correct outdated or erroneous knowledge can in
principle be used to insert false or misleading information. Our work
studies whether and how locate-then-edit transfers to masked diffusion
models, rather than proposing editing for deployment, and our
experiments use standard public benchmarks
(\textsc{CounterFact}, \textsc{Kamel}) and openly available models. We
nonetheless note that editing methods should be applied with
appropriate safeguards---such as provenance tracking and human
oversight---particularly as they extend to new model families. We see
no additional risks specific to our analysis beyond those already
present in the knowledge-editing literature.

\subsubsection*{Acknowledgments}

This work was supported by the National Research Foundation of Korea (NRF) grant (RS-2024-00333484) and the Institute of Information \& Communications Technology Planning \& Evaluation (IITP) grant (RS-2024-00338140, Development of Learning and Utilization Technology to Reflect Sustainability of Generative Language Models and Up-to-dateness over Time), both funded by the Korean government (MSIT). This work was also supported by the Institute of Information \& Communications Technology Planning \& Evaluation (IITP) grant funded by the Korean government (MND) (No. RS-2026-25551943, Military-Industry-Academia Cooperation Center (Seoul (Yongsan))), and by the National Supercomputing Center with supercomputing resources including technical support (KSC-2025-CRE-0332).

\bibliography{custom}
\newpage
\appendix
\section{MEMIT Closed-Form Update}
\label{app:memit}
This appendix expands the closed-form update summarized in
\S\ref{sec:prelim_ke}, following \citet{meng2023memit}. We use the
notation of \S\ref{sec:prelim_ke} throughout: $\mathbf{W}$ is the MLP
down-projection viewed as a linear associative memory, $\mathbf{k}$ a
key (the input to $\mathbf{W}$), and $\mathbf{v}$ its value (the MLP
output).

\paragraph{Single-layer associative memory.}
A layer storing a set of key--value pairs $\{(\mathbf{k}_i,
\mathbf{v}_i)\}_{i=1}^{n}$ holds the weights that minimize the squared
recall error,
\begin{equation}
\label{eq:memit_normal}
\mathbf{W} = \arg\min_{\widehat{\mathbf{W}}} \sum_{i=1}^{n}
\bigl\lVert \widehat{\mathbf{W}}\mathbf{k}_i - \mathbf{v}_i
\bigr\rVert^2 ,
\end{equation}
which, stacking the keys and values into matrices
$\mathbf{K} = [\mathbf{k}_1 \mid \cdots \mid \mathbf{k}_n]$ and
$\mathbf{V} = [\mathbf{v}_1 \mid \cdots \mid \mathbf{v}_n]$, is solved
by the normal equation $\mathbf{W}\mathbf{K}\mathbf{K}^\top =
\mathbf{V}\mathbf{K}^\top$. We write $\mathbf{C} =
\mathbf{K}\mathbf{K}^\top$ for the Gram matrix of the stored keys,
which is proportional to their uncentered covariance.

\paragraph{Adding a new association.}
An edit asks the same weights to additionally map a set of new keys
$\mathbf{K}_1$ to target values collected in $\mathbf{V}_1$ while
preserving the existing associations. Assuming that $\mathbf{W}$
satisfies the normal equation above, writing the updated weights as
$\mathbf{W}+\boldsymbol{\Delta}$, solving the normal equation for the
combined key set $[\mathbf{K} \mid \mathbf{K}_1]$ in block form, and
subtracting the original normal equation yields a closed-form
update~\citep{meng2023memit}:
\begin{equation}
\label{eq:memit_delta}
\boldsymbol{\Delta} = \mathbf{R}\,\mathbf{K}_1^\top
\bigl(\mathbf{C} + \mathbf{K}_1\mathbf{K}_1^\top\bigr)^{-1},
\end{equation}
where $\mathbf{R} = \mathbf{V}_1 - \mathbf{W}\mathbf{K}_1$ is the
residual of the new targets under the current weights. Because
pre-training is opaque, the keys $\mathbf{K}$ that the layer already
stores are not accessible, so $\mathbf{C}$ cannot be computed
directly. It is instead estimated as
$\lambda\,\mathbb{E}[\mathbf{k}\mathbf{k}^\top]$, where the expectation
is taken over an empirical sample of inputs to the layer and $\lambda$
is a hyperparameter balancing the new associations against the
preserved ones.

\paragraph{Optimizing the target value.}
The update in Eq.~\ref{eq:memit_delta} depends on $\mathbf{V}_1$ only
through the residual $\mathbf{R}$. For a single edited layer, the
target value is $\mathbf{v}^\ast = \mathbf{v} + \boldsymbol{\delta}$,
so $\mathbf{R}$ has $\boldsymbol{\delta}$ as its column, and
$\boldsymbol{\delta}$ is the quantity optimized in
Eq.~\ref{eq:delta_opt}: it is chosen so that adding it at the edit
location drives the model to produce the target object $o^\ast$.
Since the MLP output is added directly to the residual stream,
perturbing the value by $\boldsymbol{\delta}$ is equivalent to
perturbing the hidden state at that layer by $\boldsymbol{\delta}$;
\citet{meng2023memit} phrase the optimization in terms of the latter.
To make the edit robust across contexts, the objective averages over
prompts formed by prepending random prefixes to the rewriting prompt
$p$. For an ARM these prompts place $o^\ast$, which may span several
tokens, as the continuation to be predicted; for an MDM we instead
append $L$ \texttt{[MASK]} tokens and score $o^\ast$ at the masked
positions, as described in \S\ref{para:transfer}.

\paragraph{Spreading the update over layers.}
Rather than editing a single layer, MEMIT distributes the update over
the contiguous range $\mathcal{R}$ of critical MLP layers identified
by causal tracing. The residual $\boldsymbol{\delta}$ is optimized
once, at the hidden state of the last layer $L = \max(\mathcal{R})$,
and the layers $l \in \mathcal{R}$ are then updated in ascending
order, each with its own key matrix $\mathbf{K}_1^{l}$ and covariance
estimate $\mathbf{C}^{l}$. Let $\boldsymbol{\delta}^{(l)}$ denote the
residual that remains at layer $L$ when layer $l$ is reached, i.e.,
the gap between the optimized hidden state and the current hidden
state at layer $L$ after the layers below $l$ have been edited. The
target values for layer $l$ are set to its current outputs plus an
even share of this remaining residual,
\begin{equation}
\label{eq:memit_spread}
\mathbf{V}_1^{l} = \mathbf{W}^{l}\mathbf{K}_1^{l} + \mathbf{R}^{l},
\qquad
\mathbf{R}^{l} = \frac{1}{L - l + 1}\,
\bigl[\boldsymbol{\delta}_1^{(l)} \mid \cdots \mid
\boldsymbol{\delta}_u^{(l)}\bigr],
\end{equation}
where $L - l + 1$ is the number of layers in $\mathcal{R}$ that have
yet to be edited, including $l$, so that each of them absorbs an
approximately equal portion of the change. Because editing one layer
shifts the activations of all later ones, both the keys
$\mathbf{K}_1^{l}$ and the remaining residual
$\boldsymbol{\delta}^{(l)}$ are recomputed after every layer update,
and $\boldsymbol{\Delta}^{l}$ follows from Eq.~\ref{eq:memit_delta}
with $(\mathbf{K}_1^{l}, \mathbf{C}^{l}, \mathbf{R}^{l})$ in place of
$(\mathbf{K}_1, \mathbf{C}, \mathbf{R})$.

\section{Evaluation Details}
\label{app:eval}
We adopt the three standard knowledge-editing
criteria~\citep{meng2023locating, meng2023memit, fang2025alphaedit},
but assess all of them from the model's \emph{generation} rather than
from probability comparisons, so that the same protocol applies
uniformly to ARMs and MDMs. For each item $i$, let $x_i$ be the
rewrite prompt $(s_i, r_i)$, $N(x_i)$ its paraphrase prompts, and
$O(x_i)$ its neighborhood prompts (a distinct subject $s_i'$ sharing
the relation $r_i$ and object $o_i$). Let $o^\ast$ denote the new
target object and $o$ the original object.

\paragraph{Generation protocol.}
Given a prompt, the model generates a continuation of length
$T = \max\bigl(\lvert\tau(\sqcup o^\ast)\rvert, \lvert\tau(\sqcup o)\rvert\bigr)$,
where $\tau(\cdot)$ denotes tokenization and $\sqcup$ a leading space.
A single generation of this length is scored against both objects,
which lets us track not only whether the new target $o^\ast$ appears
but also whether the original $o$ still does. For ARMs we use greedy
left-to-right decoding of $T$ tokens. For MDMs we run denoising over a
fully masked suffix, committing one position per step in descending
order of model confidence, with the number of steps equal to the
suffix length. 
Both MDMs use a suffix of length $T$.

\paragraph{Scoring.}
An edit \emph{hits} a target $t$ on a prompt $p$ if $t$ appears as a
case-insensitive substring of the generated text (after whitespace
stripping):
\begin{equation}
\mathrm{hit}(p, t) = \mathbf{1}\bigl[\,t \subseteq_{\text{substr}}
\mathrm{gen}(p)\,\bigr].
\end{equation}
The three criteria are then
\begin{align}
\text{Efficacy} &= \mathbb{E}_i\,\mathrm{hit}(x_i, o^\ast), \\
\text{Generalization} &= \mathbb{E}_i\,\mathbb{E}_{p \in N(x_i)}\,\mathrm{hit}(p, o^\ast), \\
\text{Specificity} &= \mathbb{E}_i\,\mathbb{E}_{p \in O(x_i)}\,\mathrm{hit}(p, o),
\end{align}
where efficacy and generalization check that the new target $o^\ast$
is produced on the rewrite and paraphrase prompts, while specificity
checks that the original object $o$ is \emph{preserved} on
neighborhood prompts.

\section{Localization Analyses: Details and Additional Results}
\label{app:tracing}

\subsection{Causal Tracing}

This appendix details the causal tracing setup summarized in
\S\ref{sec:tracing}, following the procedure of
\citet{meng2023locating}.

\paragraph{Dataset and fact selection.}
Tracing is run on the \textsc{Knowns} dataset of
\citet{meng2023locating} (1{,}209 facts), separate from the
\textsc{CounterFact} and \textsc{Kamel} sets used for editing
evaluation. Following \citet{meng2023locating}, we restrict the
analysis to facts that the clean model already predicts correctly,
i.e., whose first decoded token of $o$ matches the full answer string.
Since this filter depends on each model's own predictions, the
retained subset differs across the four models.

\paragraph{MDM Adaptation.}
Because an MDM predicts at mask positions rather than at a next token,
we append a block of mask tokens to the prompt and read the score at
the first of them. The score $P(o)$ is the probability of the first
token of $o$ at this position. The correctness filter above---that the
first decoded token match the full answer string---restricts the trace
to single-token answers, exactly as in the ARM case, where the same
filter is applied to the first-token prediction at the last prompt
position. We append a block of $8$ mask tokens rather than a single
one so that the input resembles the generation-time context: because
the MDM attends bidirectionally, the number of appended masks can
affect the hidden states at the subject tokens. With single-token
answers, the answer is always scored at the first mask position.

Tracing uses a single forward pass over the masked input rather than
the iterative denoising used at evaluation. A single pass exposes each
layer's contribution directly, which is what layer-by-layer
restoration requires; iterative denoising would instead entangle
contributions across steps.

The corruption is applied exactly as in the ARM case: Gaussian noise
is added to the embeddings of the subject tokens only, leaving the
appended mask block untouched. The noise level $\nu = 3\sigma$ is
calibrated to each model, with $\sigma$ the standard deviation of
subject-token embeddings, estimated separately per model from its own
embeddings.

\paragraph{Tracing Window.}
When tracing the full hidden state we restore a single state
$h_{\hat{i}}^{(\hat{l})}$. When tracing a sub-module (MLP or
attention), a single layer's contribution is too small to recover the
prediction on its own, so we restore a contiguous window of $10$
layers centered at $\hat{l}$ at token $\hat{i}$, following
\citet{meng2023locating}.

\paragraph{Normalized AIE.}
We report the \emph{normalized AIE}, the standard fractional tracing
effect of \citet{meng2023locating}: the recovery a restored state
provides as a fraction of the gap that corruption induces,
\begin{equation}
\label{eq:norm_aie}
\resizebox{\columnwidth}{!}{$
\mathrm{normAIE}(\hat{i}, \hat{l}) =
\mathbb{E}_{\text{facts}}\!\left[
\frac{P_{\text{restored}}(o) - P_{\text{corrupted}}(o)}
     {P_{\text{clean}}(o) - P_{\text{corrupted}}(o)}
\right] \times 100\%
$}
\end{equation}
where $P_{\text{clean}}$, $P_{\text{corrupted}}$, and
$P_{\text{restored}}$ are the first-token probabilities of $o$ under
the clean, corrupted, and corrupted-with-restoration runs. A value of
$0$ means no recovery beyond the corrupted run and $100\%$ a full
recovery. Reporting it as a percentage places the four models on a
common scale despite their different baseline probabilities.

\paragraph{Within-model: confidence sharpens localization at a fixed site.}
The MDMs show a weaker average effect than the ARMs. Before comparing
the paradigms, we must ask whether this gap simply reflects
confidence, since the models differ markedly in their
$P_{\text{clean}}(o)$ distributions (Table~\ref{tab:pclean_dist}). A
comparison of peak AIE across models is only meaningful if the peak
occurs at the same place in every model: if its location on the
tracing grid (the token--layer position where restoration is most
effective) shifted with confidence, a cross-model gap in peak height
could reflect a difference in \emph{where} recall happens rather than
\emph{how concentrated} it is. We therefore first check whether the
peak location is stable. Splitting each model's traced facts into
$P_{\text{clean}}(o)$ tertiles and recomputing the normalized AIE
within each (Table~\ref{tab:tertile_peak}), we find that confidence
changes how strong the effect is, but not where it lives: in every
tertile and every model, the effect peaks at the same location, the
early-to-mid layers at the last subject token, and in every model the
top tertile's peak exceeds the bottom's
(LLaMA $72.6\!\to\!85.9$, Qwen $72.0\!\to\!77.0$, LLaDA
$17.7\!\to\!31.0$, Dream $33.6\!\to\!52.6$). Because the peak location
does not move, the matched comparison below compares all four models
at a single, shared site, and differences in peak height can be read
as differences in effect size rather than in location.

\paragraph{Across models: the MDM gap persists at matched confidence.}
The within-model result raises a concern for the cross-model
comparison: the ARMs' higher average AIE might reflect nothing more
than their having more high-confidence facts.
Table~\ref{tab:pclean_dist} already hints that this cannot be the whole
story, because LLaDA is more confident than Qwen on average
($0.54$ vs.\ $0.49$) yet shows a far lower AIE. To settle the
question, we compare the paradigms at matched confidence, binning
facts by absolute $P_{\text{clean}}(o)$
(Table~\ref{tab:matched_pclean}); facts with
$P_{\text{clean}}(o) < 0.1$ are excluded, as the normalized ratio
becomes unstable when the corruption gap is small. The result is
unambiguous. In every bin the MDM peak stays well below its paired ARM
(LLaDA reaches $18$ to $57\%$ of LLaMA, and Dream $41$ to $72\%$ of
Qwen), even though the two models' within-bin mean
$P_{\text{clean}}(o)$ is nearly identical
(Table~\ref{tab:matched_stats}). The gap is narrowest at high
confidence but never closes. The weaker MDM signal is therefore
intrinsic: factual recall in the MDMs is genuinely spread across more
layers and positions, rather than merely concentrated on
lower-confidence facts.

\begin{table}[t]
\centering
\small
\begin{tabular}{lcccc}
\toprule
Model & Mean & Median & $t_1$ & $t_2$ \\
\midrule
LLaMA-3-8B  & 0.71 & 0.78 & 0.60 & 0.91 \\
LLaDA-8B    & 0.54 & 0.56 & 0.41 & 0.70 \\
Qwen2.5-7B  & 0.49 & 0.46 & 0.35 & 0.59 \\
Dream-v0-7B & 0.45 & 0.42 & 0.28 & 0.57 \\
\bottomrule
\end{tabular}
\caption{Distribution of $P_{\text{clean}}(o)$, the clean-run
first-token probability of the correct object, over the traced facts
of each model. The tertile boundaries $t_1$ (lower) and $t_2$ (upper)
are the cut points used for the stratified analysis.}
\label{tab:pclean_dist}
\end{table}

\begin{table}[t]
\centering
\small
\begin{tabular}{lccc}
\toprule
Model & Low & Mid & High \\
\midrule
LLaMA-3-8B  & 72.6 & 65.4 & 85.9 \\
Qwen2.5-7B  & 72.0 & 68.9 & 77.0 \\
LLaDA-8B    & 17.7 & 16.7 & 31.0 \\
Dream-v0-7B & 33.6 & 34.1 & 52.6 \\
\bottomrule
\end{tabular}
\caption{Peak of the mean normalized-AIE curve (\%) at the
last-subject early site, within $P_{\text{clean}}(o)$ tertiles
(boundaries $t_1,t_2$ in Table~\ref{tab:pclean_dist}). In every model
the peak is highest in the top tertile, at an unchanged location.}
\label{tab:tertile_peak}
\end{table}

\begin{table}[t]
\centering
\small
\begin{tabular}{lcccc}
\toprule
$P_{\text{clean}}(o)$ bin & LLaMA & LLaDA & Qwen & Dream \\
\midrule
$[0.1,0.2)$ & 71 & 21 & 84 & 34 \\
$[0.2,0.3)$ & 45 & 18 & 65 & 28 \\
$[0.3,0.4)$ & 71 & 16 & 76 & 34 \\
$[0.4,0.5)$ & 78 & 14 & 77 & 35 \\
$[0.5,0.6)$ & 76 & 19 & 66 & 41 \\
$[0.6,0.7)$ & 59 & 23 & 74 & 45 \\
$[0.7,0.8)$ & 68 & 20 & 74 & 46 \\
$[0.8,0.9)$ & 66 & 38 & 85 & 49 \\
$[0.9,1.0)$ & 86 & 49 & 93 & 67 \\
\bottomrule
\end{tabular}
\caption{Peak of the mean normalized-AIE curve (\%) at the
last-subject early site, for facts binned by absolute
$P_{\text{clean}}(o)$. Within each AR/MDM pair (LLaMA/LLaDA,
Qwen/Dream) the MDM peak is lower in every bin. Ratios quoted in the
text are computed from unrounded values; per-bin counts and within-bin
mean $P_{\text{clean}}(o)$ are given in
Table~\ref{tab:matched_stats}. The dip for LLaMA in $[0.2,0.3)$
reflects bin-level variability at small $n$
($n{=}32$; Table~\ref{tab:matched_stats}) and does not affect the
pairwise comparison.}
\label{tab:matched_pclean}
\end{table}

\begin{table}[t]
\centering
\small
\setlength{\tabcolsep}{3.5pt}
\begin{tabular}{lcccccccc}
\toprule
& \multicolumn{2}{c}{LLaMA}
& \multicolumn{2}{c}{LLaDA}
& \multicolumn{2}{c}{Qwen}
& \multicolumn{2}{c}{Dream} \\
\cmidrule(lr){2-3}
\cmidrule(lr){4-5}
\cmidrule(lr){6-7}
\cmidrule(lr){8-9}
$P_{\text{clean}}(o)$ bin
& $n$ & $\bar{P}$
& $n$ & $\bar{P}$
& $n$ & $\bar{P}$
& $n$ & $\bar{P}$ \\
\midrule
$[0.1,0.2)$ &  36 & .164 &  53 & .149 &  53 & .158 & 109 & .151 \\
$[0.2,0.3)$ &  32 & .246 &  59 & .251 & 124 & .251 &  89 & .252 \\
$[0.3,0.4)$ &  62 & .352 &  52 & .343 &  93 & .348 &  76 & .352 \\
$[0.4,0.5)$ &  73 & .452 &  66 & .448 & 100 & .444 &  66 & .448 \\
$[0.5,0.6)$ &  77 & .555 &  62 & .550 &  89 & .545 &  79 & .545 \\
$[0.6,0.7)$ &  76 & .650 &  71 & .654 &  82 & .650 &  55 & .643 \\
$[0.7,0.8)$ &  76 & .752 &  57 & .747 &  49 & .751 &  52 & .751 \\
$[0.8,0.9)$ & 120 & .855 &  72 & .852 &  50 & .849 &  58 & .850 \\
$[0.9,1.0)$ & 283 & .961 &  56 & .927 &  36 & .951 &  37 & .934 \\
\bottomrule
\end{tabular}
\caption{Per-bin fact counts and within-bin mean
$P_{\text{clean}}(o)$ for the matched comparison of
Table~\ref{tab:matched_pclean}. Within each AR/MDM pair the within-bin
means agree to $\leq 0.016$ in all bins except the topmost
($\leq 0.034$).}
\label{tab:matched_stats}
\end{table}

\subsection{Validation with Additional Localization Techniques}
\label{app:additional_localization}

Section~\ref{sec:tracing} uses causal tracing to show that factual
recall in both MDMs is localized to the early-to-mid-layer MLP at the
last subject token. To test whether this conclusion depends on the
choice of localization technique, we additionally apply
gradient-based attribution and MLP knockout to LLaDA-8B and
Dream-v0-7B. We compare the layer-wise patterns identified by these
three localization techniques and also relate them to the direct
editing sweep in Section~\ref{sec:editing}.

\paragraph{Experimental setup.}
All localization analyses use the same \textsc{Knowns} source pool
and masked-input construction as Section~\ref{sec:tracing}. We append
a block of eight mask tokens to each factual prompt and analyze the
probability of the correct object at the first mask position. For
each technique, we retain facts for which the corresponding clean
forward pass predicts the correct object. The resulting fact sets
differ slightly across techniques, but recomputing the correlations
on their common subset changes the values reported below by at most
.01.

\paragraph{Gradient-based attribution.}
We apply activation-times-gradient attribution at the level of MLP
output vectors. Let $\mathbf{m}_{i}^{(\ell)} \in \mathbb{R}^{d}$ denote the MLP output at
token position $i$ and layer $\ell$. We compute
\begin{equation}
\alpha_{i}^{(\ell)}
= \sum_{h=1}^{d}
m_{i,h}^{(\ell)}\,
\frac{\partial \log P(o)}{\partial m_{i,h}^{(\ell)}} ,
\end{equation}
where $o$ is the correct object predicted at the first mask position.
A single forward and backward pass yields scores for every token
position and layer. Because our goal is token--layer localization
rather than neuron-level attribution, we operate on the full MLP
output vector using standard backpropagation. We average the absolute
attribution magnitude, $|\alpha_{i}^{(\ell)}|$, across facts to
measure localization strength without cancellation between positive
and negative contributions.

\paragraph{MLP knockout.}
At the last subject token, we jointly replace the MLP outputs in a
centered 10-layer window with zero. We measure the resulting decrease
in the log-probability of the correct object,
\begin{equation}
\Delta \log P(o)
= \log P_{\mathrm{clean}}(o)
- \log P_{\mathrm{knockout}}(o).
\end{equation}
Whereas gradient attribution provides a local first-order estimate,
MLP knockout directly intervenes on the corresponding MLP window and
therefore provides an independent validation of the localization
result.

\paragraph{Editing sweep and comparison.}
We additionally compare the localization results with the direct
editing sweep from Section~\ref{sec:editing}. The sweep moves a
4-layer editing window across layers with stride 1 at the last subject
token and measures editing efficacy. Because this experiment uses an
editing dataset rather than \textsc{Knowns}, we compare the overall
layer-wise patterns rather than individual facts.

Causal tracing and MLP knockout already operate at the same 10-layer
resolution. We aggregate the layer-level attribution scores over the
same centered 10-layer windows before comparison. When comparing the
localization results with the editing sweep, we aggregate the
localization scores over the corresponding 4-layer editing windows.
Agreement is measured using Spearman rank correlation across layer
windows. For visualization, each curve is independently min--max
normalized; the correlations are computed from the original scores
after matching their window resolutions.

\begin{figure*}[t]
    \centering
    \includegraphics[width=\textwidth]
    {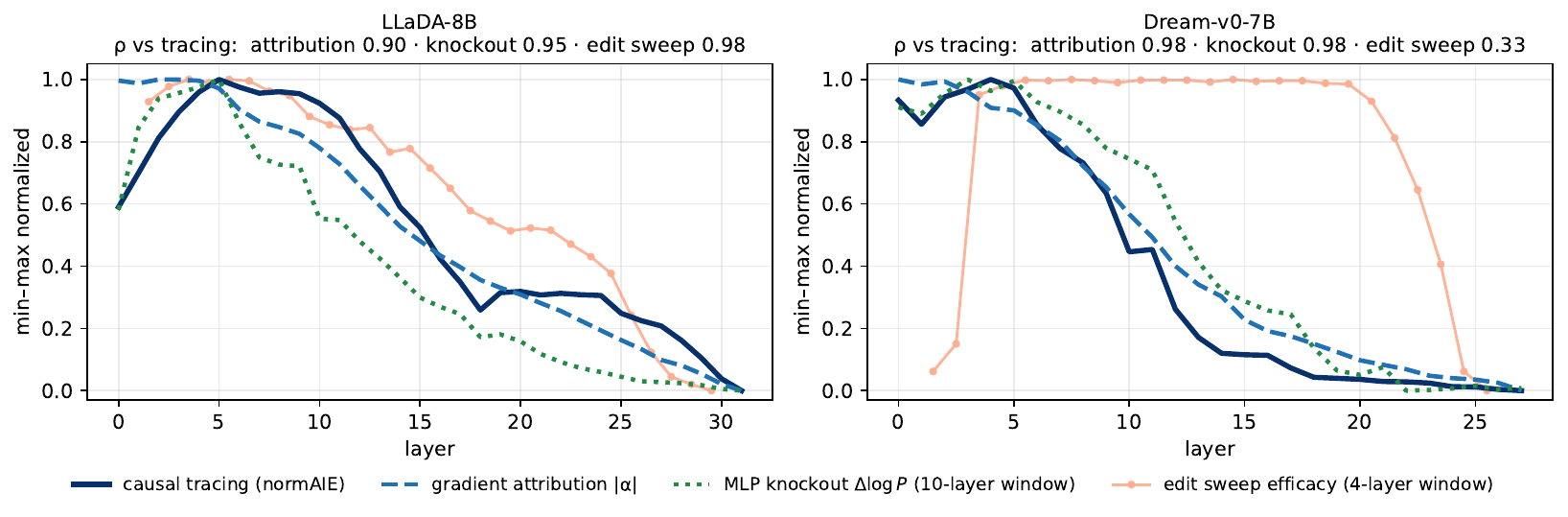}
    \caption{
    Layer-wise comparison of causal tracing, gradient-based
    attribution, MLP knockout, and the direct editing sweep at the
    last subject token. Curves are independently min--max normalized;
    $\rho$ denotes Spearman correlation after matching window
    resolutions.
    }
    \label{fig:additional_localization}
\end{figure*}

\paragraph{Results.}
Gradient-based attribution and MLP knockout closely agree with causal
tracing in both models. The rank correlation between attribution and
tracing is $\rho=.90$ for LLaDA and $\rho=.98$ for Dream, while the
correlation between knockout and tracing is $\rho=.95$ and $\rho=.98$,
respectively. Gradient attribution is strongest at the last subject
token, and at this position both attribution and knockout concentrate
in the early-to-mid layers before decreasing toward later layers.
Thus, intervention-based tracing, gradient-based attribution, and
ablation-based knockout all support the same token--layer location.
The localization result in Section~\ref{sec:tracing} is therefore not
specific to causal tracing.

The relationship between localization and the most effective editing
layers is model-dependent. For LLaDA, causal tracing and the editing
sweep vary similarly across layers ($\rho=.98$). For Dream, however,
the localization scores decrease after the early-to-mid layers,
whereas editing efficacy remains high across a substantially broader
layer range, yielding a lower correlation ($\rho=.33$). This does not
reflect disagreement among the localization techniques, which remain
strongly aligned with one another on Dream. Instead, it is consistent
with the observation that the components most causally involved in
recall need not be the most effective locations to edit
\citep{hase2023}.

Causal localization therefore provides evidence about where factual
recall occurs, but does not by itself determine the optimal editing
layers. This motivates our choice in Section~\ref{sec:editing} to
measure editing performance directly across token positions and layer
windows rather than selecting the editing site solely from the causal
tracing result.

\section{Locate-then-Edit on MDMs}
\label{app:editing}
This appendix expands the transfer described in \S\ref{para:transfer}
to the level of the implementation. The closed-form solver, the
key/value framework, and the delta optimization loop of MEMIT
\citep{meng2023memit} are kept intact; the only change is that every
input distribution on which MEMIT runs a forward pass is
mask-augmented to match the bidirectional attention of an MDM.

A single MEMIT edit runs forward passes on four distinct input
distributions. In the ARM case all of them use the raw prompt (or
prompt followed by the target); in the MDM case we put all of them in
the same form, a prompt followed by a block of mask tokens at the
answer slot, with one mask token per token of the edit target
$o^\ast$.
\begin{enumerate}
  \item \textbf{Rewriting prompts.} The main term of
  Eq.~\ref{eq:delta_opt}: we append the mask block to each fact's
  rewriting prompt and supervise the logits at the mask positions
  toward $o^\ast$. The ARM's next-token logits at the final answer
  positions thus map onto the logits read at the mask positions.
  \item \textbf{KL-anchor prompt.} Beyond the term shown in
  Eq.~\ref{eq:delta_opt}, the optimization also includes a KL
  regularizer that keeps unrelated predictions stable. Its anchor is
  likewise mask-augmented: in place of the standard subject-last
  next-token logits, we read the logits at the mask position of the
  anchor prompt followed by a mask token.
  \item \textbf{Key extraction.} The key $\mathbf{k}$ for each fact is
  read as the subject-last hidden state on the mask-augmented prompt.
  \item \textbf{Residual baseline.} The baseline value used to form
  the closed-form residual is likewise read from the subject-last
  hidden state on the mask-augmented prompt.
\end{enumerate}
Under the causal attention of an ARM, the subject-last hidden state is
identical on the raw prompt and on the prompt followed by the target,
since future tokens cannot be attended to; the two hidden states that
MEMIT reads---the optimization target and the residual baseline---then
coincide automatically. Under the bidirectional attention of an MDM
this coincidence breaks, so we force all four distributions to be
defined over the same masked context.

\paragraph{Hyperparameters.}
The two ARMs follow the MEMIT settings provided in
EasyEdit~\citep{wang-etal-2024-easyedit}: learning rate $10^{-1}$, 
clamp-norm factor $0.75$, and KL
factor $0.0625$. For the two MDMs, we inherit the hyperparameters of
their architectural sibling without retuning: LLaDA from LLaMA (both
share the LLaMA-style transformer block) and Dream from Qwen (Dream's
backbone is Qwen2.5). The closed-form solver, the covariance term, the
weight update, and the distribution of the residual across layers are
all as in the original MEMIT~\citep{meng2023memit}. The only setting
we determine per model is the \emph{edit layer}, chosen by a direct
layer sweep that selects the best $4$-layer window by the rank-sum of
efficacy and generalization; this yields LLaMA-3 L$1$--$4$, LLaDA
L$4$--$7$, Qwen2.5 L$7$--$10$, and Dream L$4$--$7$
(Appendix~\ref{app:layer_selection}).

\section{Editing Layer Selection}
\label{app:layer_selection}
The edit layer is the one setting we tune per model, since editing
performance is sensitive to which MLP layers are modified and prior
defaults do not transfer reliably across our four models. For each
model we sweep every $4$-layer contiguous window on a held-out subset
of \textsc{CounterFact} ($500$ facts), score each window by the
rank-sum of its efficacy and generalization ranks, and select the
best-scoring window (ties broken by efficacy). We exclude specificity
from the criterion because it trades off against efficacy: the
highest-specificity windows tend to under-edit. The sweep uses the
same evaluation settings as the main experiments
(Appendix~\ref{app:eval}) and the hyperparameters of
Appendix~\ref{app:editing}.

\section{ARM Editing Results}
\label{app:arm_editing}

\begin{figure}[h]
    \centering
        \includegraphics[width=\linewidth]{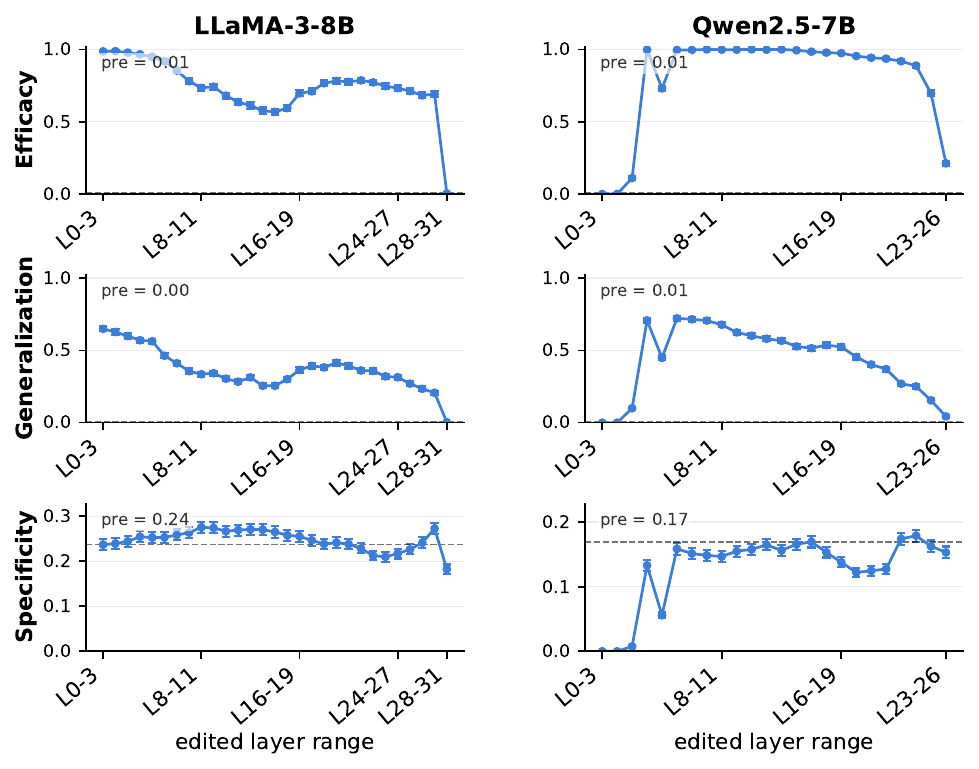}
            \caption{Layer sweep for ARM editing.}
                \label{fig:layer_sweep_arm}
                \end{figure}

\begin{figure}[h]
                    \centering
                        \includegraphics[width=\linewidth]{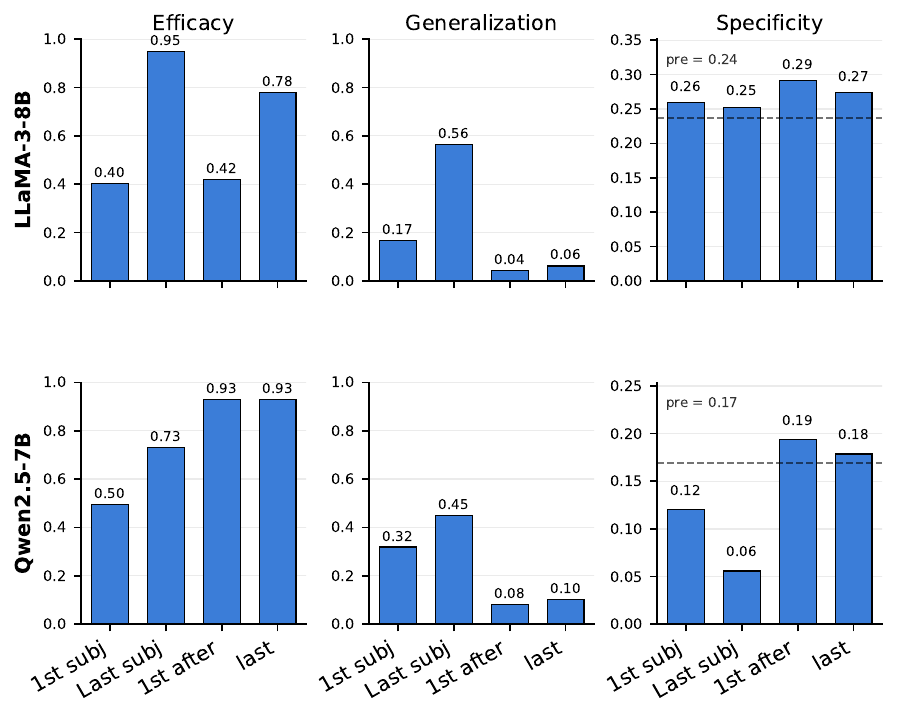}
                            \caption{Token position ablation for ARM editing.}
                                \label{fig:token_position_arm}
                                \end{figure}

For completeness, we report the ARM counterparts of the editing
analyses in the main text. Figure~\ref{fig:layer_sweep_arm} shows the
layer sweep for the two ARMs, and Figure~\ref{fig:token_position_arm}
shows the token position ablation. The trends mirror those of the
MDMs in \S\ref{sec:editing}.

\section{Robustness to Generation Settings}
\label{app:generation_robustness}

Table~\ref{tab:gen_robust} reports the full results of the
generation-robustness analysis in \S\ref{sec:editing-robust}. We
evaluate each edited model without further optimization while varying
the generation length and number of denoising steps.

\begin{table}[t]
\centering
\small
\setlength{\tabcolsep}{4pt}
\begin{tabular}{l ccc ccc}
\toprule
& \multicolumn{3}{c}{LLaDA}
& \multicolumn{3}{c}{Dream} \\
\cmidrule(lr){2-4}
\cmidrule(lr){5-7}
Length / steps
& Eff. & Gen. & Spec.
& Eff. & Gen. & Spec. \\
\midrule
pre-edit      & 0.01 & 0.01 & 0.18 & 0.00 & 0.00 & 0.17 \\
target-length & 0.91 & 0.56 & 0.19 & 0.97 & 0.51 & 0.20 \\
8 / 8         & 0.82 & 0.45 & 0.28 & 0.95 & 0.56 & 0.30 \\
16 / 16       & 0.86 & 0.51 & 0.27 & 0.94 & 0.59 & 0.34 \\
32 / 32       & 0.84 & 0.50 & 0.25 & 0.93 & 0.59 & 0.35 \\
\bottomrule
\end{tabular}
\caption{
Editing performance on \textsc{CounterFact} under varying generation
lengths and numbers of denoising steps. The dataset contains mostly
single-token targets. \emph{Target-length} matches the generation
length to each edit target, and \emph{pre-edit} reports the unedited
model under the same setting.
}
\label{tab:gen_robust}
\end{table}

\section{Additional Analysis of Multi-Token Editing Failures}
\label{app:length_failure}

\paragraph{Target-length trend on \textsc{CounterFact}.}
The controlled \textsc{Kamel} study in
\S\ref{sec:failure-degrade} was motivated by a length trend already
visible on \textsc{CounterFact}. Here we report the per-length editing
results on the \textsc{CounterFact} sweep set ($n=500$), with each
model evaluated at its selected edit layer
(Appendix~\ref{app:layer_selection}). Because \textsc{CounterFact} is
single-token dominated, the multi-token groups contain few facts; we
report them to show the trend rather than as precise estimates.

\begin{table}[h]
\centering
\small
\setlength{\tabcolsep}{6pt}
\begin{tabular}{lcc}
\toprule
Target length & LLaDA & LLaMA \tabularnewline
\midrule
$N=1$   & $0.92$ ($443/481$) & $0.99$ ($482/487$) \tabularnewline
$N=2$   & $0.58$ ($7/12$)    & $0.92$ ($12/13$)   \tabularnewline
$N=3$   & $0.57$ ($4/7$)     & ---                \tabularnewline
Overall & $0.91$ ($454/500$) & $0.99$ ($494/500$) \tabularnewline
\midrule
        & Dream              & Qwen               \tabularnewline
\midrule
$N=1$   & $0.98$ ($476/487$) & $1.00$ ($487/487$) \tabularnewline
$N=2$   & $0.77$ ($10/13$)   & $1.00$ ($13/13$)   \tabularnewline
Overall & $0.97$ ($486/500$) & $1.00$ ($500/500$) \tabularnewline
\bottomrule
\end{tabular}
\caption{Editing efficacy on the \textsc{CounterFact} sweep set
($n=500$), reported as a fraction in $[0,1]$ and broken down by target
token length under each model's tokenizer, grouped by architectural
pair (top: LLaDA/LLaMA; bottom: Dream/Qwen). Counts are
successes/total; lengths absent under a tokenizer are dashed or
omitted. Both MDMs drop sharply on multi-token targets, whereas the
paired ARMs lose only ${\sim}0.07$ (LLaMA) or nothing (Qwen) from
$N=1$ to $N=2$.}
\label{tab:cf_length_failure}
\end{table}

The single-token gap between each MDM and its ARM sibling is modest
($0.02$ for Dream vs.\ Qwen and $0.07$ for LLaDA vs.\ LLaMA), but it
widens sharply with target length. At $N=2$, it reaches $0.23$ for
Dream and $0.34$ for LLaDA, while the ARMs themselves lose only
$0.07$ (LLaMA) or nothing (Qwen) relative to $N=1$. The same effect
appears on \textsc{Kamel} under controlled length groups
(\S\ref{sec:failure-degrade}), where the larger per-length samples
make the trend precise. The agreement between the two datasets
indicates that the effect is not an artifact of \textsc{CounterFact}.

\paragraph{How intermediate-state failures appear in generation.}
The controlled state probe in \S\ref{sec:failure-diagnose} shows that
an edit optimized on the fully masked state can weaken after target
tokens are revealed. We further examine how this instability appears
in final generations. Specifically, we compare whether individual
target tokens appear anywhere in the generation with whether the
complete target is produced.

\begin{table}[h]
\centering
\small
\setlength{\tabcolsep}{5pt}
\begin{tabular}{ccccc}
\toprule
$N$ & \#Facts & Token presence & Editing efficacy & Gap (pp)
\tabularnewline
\midrule
2 & 200 & 73.5\% & 59.5\% & 14.0 \tabularnewline
3 & 200 & 53.8\% & 32.5\% & 21.3 \tabularnewline
4 & 200 & 50.7\% & 26.5\% & 24.2 \tabularnewline
\bottomrule
\end{tabular}
\caption{Token presence and editing efficacy for LLaDA on multi-token
\textsc{Kamel} targets. Token presence is averaged over target-token
positions, counting a token as present if it appears anywhere in the
length-matched generation, irrespective of position or order. Editing
efficacy requires the complete target to be generated.}
\label{tab:position_vs_joint}
\end{table}

Token presence is consistently higher than editing efficacy, and the
gap widens with target length. Thus, failed generations often retain
some constituent target tokens without producing the complete target.
This output-level pattern is consistent with the state-level finding
that the edit does not remain equally effective throughout iterative
generation.

Representative examples illustrate the same pattern. For the target
``trance'', tokenized as [``tr'', ``ance''], the model may preserve
only the second token and generate ``dance'', or preserve only the
first and generate ``trit''.

\begin{table}[h]
\centering
\small
\begin{tabularx}{\columnwidth}{XXl}
\toprule
Target & Generated & Pattern \tabularnewline
\midrule
\makecell[l]{\textit{trance}\\= [tr, ance]}
&
\makecell[l]{\textit{dance}\\= [d, ance]}
&
2nd correct, 1st wrong
\tabularnewline
\makecell[l]{\textit{trance}\\= [tr, ance]}
&
\makecell[l]{\textit{trit}\\= [tr, it]}
&
1st correct, 2nd wrong
\tabularnewline
\bottomrule
\end{tabularx}
\caption{Representative multi-token editing failures in the MDM.
The generation retains one constituent target token but fails to
produce the complete target.}
\label{tab:failure_cases}
\end{table}

Together, these output-level analyses complement the direct
intermediate-state probe in \S\ref{sec:failure-diagnose}: target
information is often partially retained, but it is not consistently
preserved across the iterative generation process.

\section{Curating \textsc{Kamel} for Multi-Token Editing}
\label{app:kamel}
We build the multi-token editing set from \textsc{Kamel}
\citep{kalo2022kamel}, recast into the \textsc{CounterFact} schema so
it runs through the same evaluation pipeline (\S\ref{sec:editing}). To
trace editing across target length we need facts at each length $N$,
which \textsc{CounterFact} does not provide; \textsc{Kamel} spans a
wide range of entity lengths. We cover $N \in \{1,2,3,4\}$ and keep a
fact at length $N$ only if its object tokenizes to exactly $N$ tokens
under \emph{both} the LLaDA and LLaMA tokenizers, so that any
ARM--MDM difference reflects target length rather than one tokenizer
splitting the same target more finely than the other.

\begin{table}[h]
\centering
\small
\setlength{\tabcolsep}{5pt}
\begin{tabular}{l rr rr}
\toprule
& \multicolumn{2}{c}{\textsc{CounterFact}} & \multicolumn{2}{c}{\textsc{Kamel}} \\
\cmidrule(lr){2-3} \cmidrule(lr){4-5}
Target length & Qwen & Dream & Qwen & Dream \\
\midrule
1 token   & 20{,}534 & 20{,}534 &  8{,}682 &  8{,}682 \\
2 tokens  &      343 &      343 & 11{,}834 & 11{,}834 \\
3 tokens  &        0 &        0 & 10{,}896 & 10{,}896 \\
4+ tokens &        0 &        0 & 21{,}875 & 21{,}875 \\
\midrule
Total     & 20{,}877 & 20{,}877 & 53{,}287 & 53{,}287 \\
\bottomrule
\end{tabular}
\caption{Distribution of target token lengths under Qwen and Dream
tokenizers (extension of Table~\ref{tab:length_dist}). The two
columns within each dataset are identical because Dream is built on
the Qwen2.5 backbone and shares its tokenizer; both confirm the same
overall picture observed under LLaMA and LLaDA --- CounterFact is
dominated by single-token targets with no facts at $N \geq 4$, while
KAMEL spans the full range with the majority of facts at $N \geq 2$.}
\label{tab:length_dist_qwen_dream}
\end{table}

\paragraph{Construction.}
Starting from the single-answer facts in \textsc{Kamel} ($54$
relations, one cloze template each), we measure object token length in
context as in our evaluation (Appendix~\ref{app:eval}) and apply the
two-tokenizer filter above. For each fact we form a counterfactual
target by swapping in another object from the same relation that is
also $N$ tokens under both tokenizers, which keeps the mask budget
fixed at $N$ and the substitution semantically plausible. Since
\textsc{Kamel} provides only one template per relation, we write one
paraphrase per relation to measure generalization. From the facts
passing all filters we sample $200$ per $N$.

\begin{table}[h]
\centering
\small
\begin{tabular}{c cccc}
\toprule
$N$ & 1 & 2 & 3 & 4 \\
\midrule
Samples           & 200 & 200 & 200 & 200 \\
Relations covered & 18  & 34  & 35  & 34  \\
\bottomrule
\end{tabular}
\caption{\textsc{Kamel} editing set by target length. The lower
relation count at $N{=}1$ reflects that objects that are single tokens
under both tokenizers concentrate in a few relations (e.g., short
country names); $N\geq2$ draws on over thirty relations.}
\label{tab:kamel_stats}
\end{table}

\section{Partial-Mask Augmentation: Details and Ablations}
\label{app:augmentation}

This appendix specifies the partial-mask augmentation of
\S\ref{sec:failure-fix} and ablates its two design choices: how
optimization steps are allocated across reveal counts, and how the
revealed positions are selected within each count.

\subsection{Method}

The correction broadens the states on which the residual
$\boldsymbol{\delta}$ (Eq.~\ref{eq:delta_opt}) is fit, leaving
everything else in MEMIT unchanged. For a target of $N$ tokens, the
first optimization step uses the fully masked configuration
($k{=}0$). At each subsequent step, we draw the number of revealed
positions $k$ uniformly at random from
$\{0,1,\dots,N{-}1\}$. Given $k$, we uniformly sample a subset of $k$
positions, fill them with the corresponding target tokens, mask the
rest, and update $\boldsymbol{\delta}$ on the $N{-}k$ still-masked
positions.

After the forced first step, each reveal count receives a $1/N$ share
of the remaining optimization steps in expectation, while resampling
the positions exposes $\boldsymbol{\delta}$ to diverse conditioning
configurations within each count. This covers different stages of
generation without assuming a fixed reveal order. Because optimization
is now distributed across $N$ reveal-count levels rather than
concentrated on the fully masked state, we enlarge the optimization
budget relative to the single-state baseline---fourfold for LLaDA and
twofold for Dream---so that each level receives sufficient
optimization; for Dream, we also loosen the norm clamp on
$\boldsymbol{\delta}$ accordingly.

\subsection{Ablation 1: Which states to visit}
\label{app:abl-kdist}
We fix everything except how the optimization steps are allocated
across reveal-count levels---that is, how often each count $k$ of
revealed positions is visited during fitting
(Table~\ref{tab:abl-kdist}, LLaDA on \textsc{Kamel}). All variants,
including the unaugmented control (a), share the enlarged budget
described above; (a) therefore differs from MEMIT$^\dagger$ in
Table~\ref{tab:correction}, which uses the default budget. This
comparison isolates the effect of the reveal-count distribution from
that of additional optimization steps.
\begin{table}[h]
\centering
\small
\setlength{\tabcolsep}{5pt}
\begin{tabular}{l ccc}
\toprule
Reveal-count distribution & $N{=}2$ & $N{=}3$ & $N{=}4$ \\
\midrule
(a) fully masked only & 0.64 & 0.44 & 0.39 \\
(b) bias toward fewer revealed & 0.88 & 0.73 & 0.69 \\
(c) bias toward more revealed & 0.85 & 0.65 & 0.55 \\
(d) bias toward intermediate counts & --- & --- & 0.69 \\
(e) deterministic cycle over counts & 0.87 & 0.76 & 0.73 \\
\midrule
\textbf{(f) uniform (ours)} & \textbf{0.87} & \textbf{0.76} & \textbf{0.73} \\
\bottomrule
\end{tabular}
\caption{Reveal-count distribution ablation (LLaDA,
\textsc{Kamel}, $n{=}200$). (a) trains only on the fully masked state
(no augmentation); (b) and (c) skew the count toward fewer or more
revealed positions; (d) skews the count toward intermediate values
($k{=}1,2$ receive 70\% of the steps; evaluated only for $N{=}4$);
(e) visits every count in a fixed deterministic cycle; and (f), our
default, draws each count uniformly at random.}
\label{tab:abl-kdist}
\end{table}
Every partial-mask-augmented variant improves over the fully masked
baseline (a), with gains ranging from 16.0 to 34.0 percentage points
(pp) across the tested settings. Our default uniform variant (f) gains
23.0, 32.0, and 34.0 pp at $N{=}2$, $N{=}3$, and $N{=}4$,
respectively. No single reveal-count distribution performs best at
every length: emphasizing fewer revealed positions (b) is strongest
at $N{=}2$, while uniform sampling (f) and deterministic cycling (e)
tie at the reported precision for $N{=}3$ and $N{=}4$. Emphasizing
fewer revealed positions consistently outperforms emphasizing more
revealed positions (c), and concentrating on intermediate counts (d)
does not improve over the equal-count schemes at $N{=}4$. We retain
random uniform sampling as our default because it covers reveal
counts without imposing a fixed schedule while matching the strongest
results at longer target lengths.

\subsection{Ablation 2: How to choose revealed positions}
\label{app:abl-reveal}

Fixing the count distribution above, we compare three ways of
choosing which $k$ positions to reveal at each step
(Table~\ref{tab:abl-reveal}, \textsc{Kamel} $N{=}3$).

\begin{table}[h]
\centering
\small
\setlength{\tabcolsep}{5pt}
\begin{tabular}{l cc cc}
\toprule
& \multicolumn{2}{c}{LLaDA} & \multicolumn{2}{c}{Dream} \\
Reveal policy & Eff. & Gen. & Eff. & Gen. \\
\midrule
left-to-right & 0.585 & 0.275 & 0.845 & 0.565 \\
by confidence & 0.615 & 0.340 & 0.815 & 0.595 \\
\midrule
\textbf{random (ours)} & \textbf{0.760} & \textbf{0.360} & \textbf{0.895} & \textbf{0.680} \\
\bottomrule
\end{tabular}
\caption{Reveal-policy ablation (\textsc{Kamel} $N{=}3$, $n{=}200$).
\emph{Left-to-right} reveals the first $k$ positions; \emph{by
confidence} reveals the $k$ positions the unedited model already
scores highest; \emph{random} resamples $k$ positions at each step.
Best per column in bold.}
\label{tab:abl-reveal}
\end{table}

Random reveal is best for both models: it improves efficacy over the next-best policy by over $14$\,pp on LLaDA and $5$\,pp on Dream, with generalization likewise highest. Resampling the revealed set each step appears to make $\boldsymbol{\delta}$ robust to many position combinations rather than to one fixed pattern, whereas the fixed left-to-right order generalizes worst on both models, suggesting $\boldsymbol{\delta}$ overfits to that specific order. We use random reveal for both models.

\section{Generality Across Models: Additional ARM--MDM Pairs}
\label{app:generality_models}

To test whether our main findings extend across model families and
scales, we evaluate three additional ARM--MDM pairs:
LLaMA-3-8B-Instruct/LLaDA-1.5-8B,
LLaMA-2-7B/DiffuLLaMA-7B, and
GPT-2-medium/DiffuGPT-m (355M). DiffuLLaMA and DiffuGPT-m are
initialized from their paired ARMs and adapted to masked diffusion,
whereas LLaDA-1.5 applies additional alignment to LLaDA-8B. Unless
otherwise noted, we follow the corresponding experimental settings
from the main text. We report additional results for causal tracing
(\S\ref{sec:tracing}), editing-site selection
(\S\ref{sec:editing}), target-length degradation
(\S\ref{sec:failure-degrade}), and intermediate-state correction
(\S\ref{sec:failure-fix}).

\subsection{Locating Facts via Causal Tracing}
\label{app:additional-tracing}

\begin{figure*}[t]
\centering
\includegraphics[width=\textwidth]{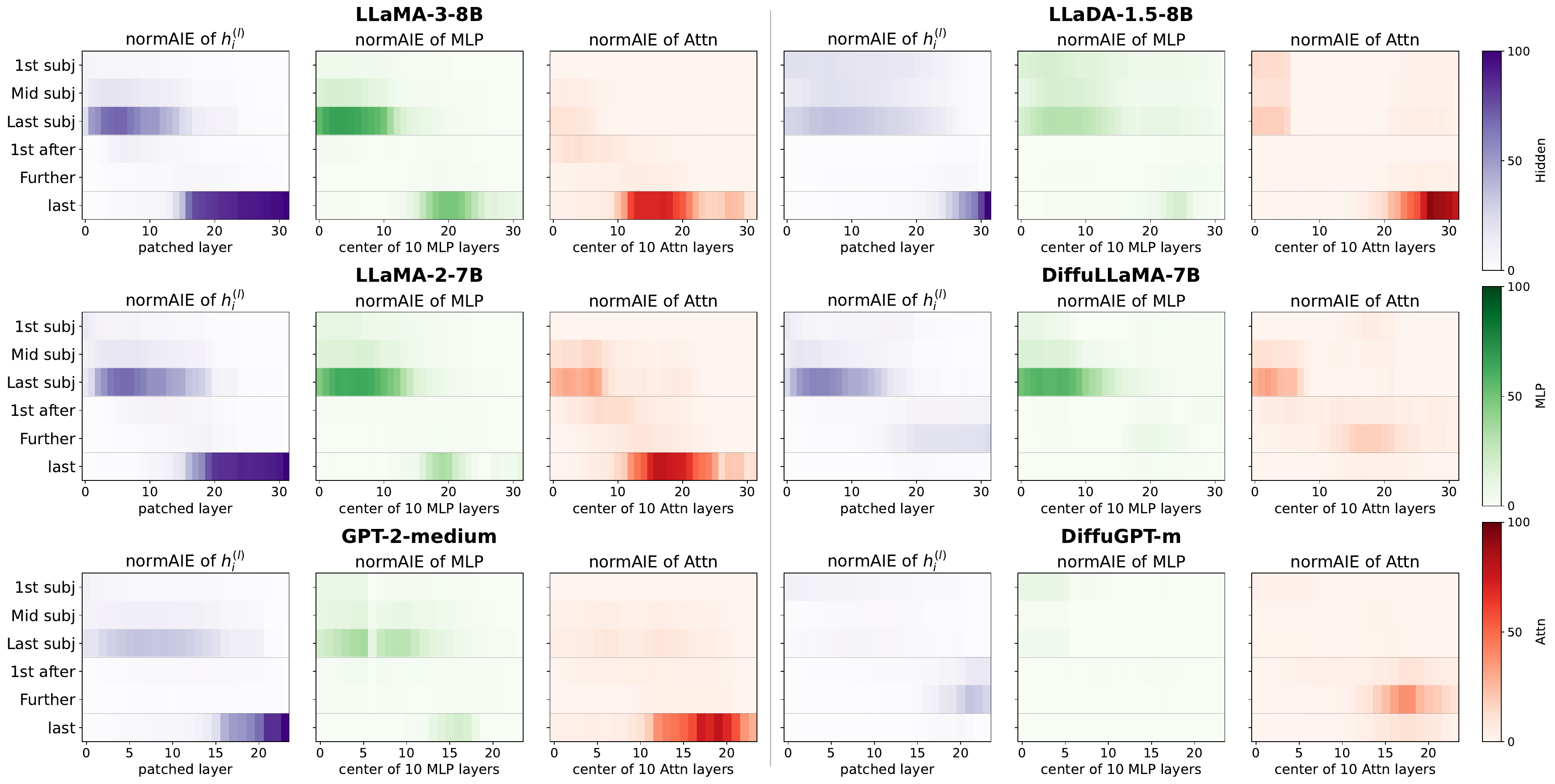}
\caption{Causal tracing across three additional ARM--MDM pairs.
Each row compares an ARM (left) with its paired MDM (right). For each
model, we report the normalized AIE from restoring the hidden state,
MLP contribution, and attention contribution across token positions
and layers. All panels use the same color scale. \texttt{last} denotes
the final input position, corresponding to the appended answer mask in
the MDMs.}
\label{fig:additional-tracing}
\end{figure*}

We apply the causal-tracing procedure from \S\ref{sec:tracing} to all
three additional ARM--MDM pairs. For each fact, we corrupt the subject
token embeddings and restore the clean hidden state, MLP contribution,
or attention contribution at different token positions and layers.
Figure~\ref{fig:additional-tracing} reports the resulting normalized
AIE.

Across the additional models, the early-layer MLP effect associated
with factual recall remains concentrated within the subject span and
decreases substantially in later layers and post-subject positions.
Five of the six models exhibit the strongest effect at the last
subject token. DiffuGPT-m instead peaks at the first subject token, but
the effect remains localized within the subject span in the early
layers. Thus, the exact peak within the subject may vary across models,
while the broader localization of the factual MLP signal to
subject-internal early layers remains consistent across model families
and scales.

\subsection{Editing-Site Selection}
\label{app:additional-editing-sites}

We next identify effective editing sites in the additional MDMs by
varying the edited layer range and token position, following
\S\ref{sec:editing}. We first fix the edited position to the last
subject token and evaluate five four-layer windows with a stride of
two on 100 \textsc{CounterFact} facts. As shown in
Table~\ref{tab:additional-layer-selection}, all three models achieve
their strongest efficacy and generalization in early-to-middle layer
windows. We therefore select layers 4--7 for LLaDA-1.5, 2--5 for
DiffuLLaMA, and 6--9 for DiffuGPT-m.

\begin{table}[t]
\centering
\small
\setlength{\tabcolsep}{3pt}
\begin{tabular}{@{}lccccc@{}}
\toprule
Model & L0--3 & L2--5 & L4--7 & L6--9 & L8--11
\tabularnewline
\midrule
LLaDA-1.5
& .89/.48 & .95/.57 & \textbf{.95/.59} & .95/.56 & .94/.53
\tabularnewline
DiffuLLaMA
& .93/.57 & \textbf{.96/.67} & .94/.58 & .91/.47 & .86/.42
\tabularnewline
DiffuGPT-m
& .81/.41 & .81/.46 & .84/.51 & \textbf{.91/.52} & .89/.39
\tabularnewline
\bottomrule
\end{tabular}
\caption{Editing-layer selection on additional MDMs using 100
\textsc{CounterFact} facts. Each entry reports
efficacy/generalization with the edited position fixed to the last
subject token. Bold indicates the selected four-layer window.}
\label{tab:additional-layer-selection}
\end{table}

Using the selected layer window for each model, we then vary the
edited token position among the first subject token, last subject
token, first post-subject token, and final input position. The last
subject token yields the highest efficacy for all three models
(Table~\ref{tab:additional-position-selection}), while efficacy
generally decreases outside the subject span.

\begin{table}[t]
\centering
\small
\setlength{\tabcolsep}{5pt}
\begin{tabular}{@{}lcccc@{}}
\toprule
Model & 1st subj. & Last subj. & 1st after & Last
\tabularnewline
\midrule
LLaDA-1.5
& .68 & \textbf{.95} & .41 & .52
\tabularnewline
DiffuLLaMA
& .65 & \textbf{.96} & .33 & .00
\tabularnewline
DiffuGPT-m
& .62 & \textbf{.91} & .65 & .25
\tabularnewline
\bottomrule
\end{tabular}
\caption{Token-position ablation on additional MDMs using their
selected layer windows. Each entry reports editing efficacy.
\texttt{Last} denotes the final input position, corresponding to the
appended answer mask.}
\label{tab:additional-position-selection}
\end{table}

These results extend the main finding that the last subject token in
the early-to-middle layers is an effective editing site. However,
DiffuGPT-m exhibits its strongest causal-tracing effect at the first
subject token while remaining most effectively edited at the last
subject token. We therefore do not interpret the precise causal peak
as necessarily identifying the optimal editing position. The two
analyses instead agree at the broader level of a subject-internal,
early-layer region associated with factual recall and editing.

\subsection{Editing Degrades with Target Length}
\label{app:additional-length-degradation}

We evaluate editing efficacy as a function of target length using the
\textsc{Kamel} setup from \S\ref{sec:failure-degrade}. For each pair,
we retain facts whose targets have the same token length under both
tokenizers and evaluate 200 facts at each length.
Table~\ref{tab:additional-length} reports the results for the three
additional pairs.

\begin{table}[t]
\centering
\small
\setlength{\tabcolsep}{3.5pt}
\begin{tabular}{@{}lcccc@{}}
\toprule
ARM / MDM & $N{=}1$ & $N{=}2$ & $N{=}3$ & $N{=}4$
\tabularnewline
\midrule
LLaMA-3 / LLaDA-1.5
& .98/.95 & .95/.60 & .92/.34 & .89/.27
\tabularnewline
LLaMA-2 / DiffuLLaMA
& .92/.99 & .87/.65 & .87/.54 & .79/.52
\tabularnewline
GPT-2-m / DiffuGPT-m
& 1.00/.86 & .90/.45 & .87/.18 & .82/.13
\tabularnewline
\bottomrule
\end{tabular}
\caption{Editing efficacy by target length on additional ARM--MDM
pairs. Each entry reports ARM/MDM efficacy on the same set of facts
($n{=}200$ per target length and pair).}
\label{tab:additional-length}
\end{table}

All three pairs achieve high efficacy on single-token targets, but the
gap between the ARM and MDM widens with target length. From $N{=}1$ to
$N{=}4$, ARM efficacy decreases by .09--.18, whereas MDM efficacy
decreases by .47--.73. The same pattern appears in DiffuLLaMA and
DiffuGPT-m despite their initialization from the corresponding ARM
weights. The multi-token degradation observed in the main text is
therefore not limited to a particular model family, initialization,
or scale and recurs across masked-diffusion models.

\subsection{A Correction for Intermediate States}
\label{app:additional-correction}

Finally, we apply the intermediate-state correction from
\S\ref{sec:failure-fix} to the three additional MDMs.
MEMIT$^\dagger$ optimizes the edit residual using only the fully masked
answer state, whereas +Corr. also includes partially unmasked states
encountered during generation. We otherwise keep the editing and
evaluation procedures unchanged and evaluate 200 facts at each target
length.

\begin{table}[t]
\centering
\small
\setlength{\tabcolsep}{4pt}
\begin{tabular}{@{}llccc@{}}
\toprule
MDM & Method & $N{=}2$ & $N{=}3$ & $N{=}4$
\tabularnewline
\midrule
LLaDA-1.5
& MEMIT$^\dagger$ & .60/.33 & .34/.20 & .27/.17
\tabularnewline
LLaDA-1.5
& +Corr. & .91/.53 & .75/.44 & .75/.42
\tabularnewline
DiffuLLaMA
& MEMIT$^\dagger$ & .65/.47 & .54/.24 & .52/.32
\tabularnewline
DiffuLLaMA
& +Corr. & .87/.50 & .78/.42 & .71/.51
\tabularnewline
DiffuGPT-m
& MEMIT$^\dagger$ & .45/.16 & .18/.07 & .13/.04
\tabularnewline
DiffuGPT-m
& +Corr. & .71/.27 & .29/.09 & .37/.08
\tabularnewline
\bottomrule
\end{tabular}
\caption{Intermediate-state correction on additional MDMs.
Each entry reports efficacy/generalization
($n{=}200$ per target length).}
\label{tab:additional-correction}
\end{table}

The correction improves both efficacy and generalization at every
target length on all three models. At $N{=}4$, efficacy increases from
.27 to .75 on LLaDA-1.5, from .52 to .71 on DiffuLLaMA, and from .13
to .37 on DiffuGPT-m. Although the magnitude of recovery varies, the
effect is consistent even for the substantially smaller DiffuGPT-m.

These results show that the mismatch between optimization on a single
fully masked state and the intermediate states visited during
iterative denoising is not specific to a particular MDM. Including
intermediate states during optimization provides a broadly effective
correction. Nevertheless, performance gaps remain for longer targets,
indicating that this mismatch is not the sole source of multi-token
editing degradation.

\section{Broader Evaluation of Edited Models}
\label{app:locality}

We more broadly evaluate models edited using MEMIT$^\dagger$, with and
without the intermediate-state correction, along three complementary
dimensions: general capability on unrelated benchmarks, generation
quality and specificity as the number of edits increases, and both the
propagation and preservation of knowledge related to the edited facts.

\subsection{General-Capability Preservation}
\label{app:locality-general}

We evaluate SST-2, MRPC, CoLA, RTE, NLI, and 5-shot MMLU, following the
general-capability evaluation used in prior editing work. We use the
same evaluation examples before and after editing and across all
conditions. We report benchmark accuracy rather than perplexity because
an MDM's likelihood is not exactly tractable and is instead estimated
through a Monte Carlo evidence lower bound over masking patterns
\citep{nie2025llada}. Such estimates introduce unnecessary variance
into fine-grained before-and-after comparisons. All six benchmarks use
single-token answers, allowing masked-position scoring consistent with
the official LLaDA evaluation protocol.

\begin{table*}[t]
\centering
\small
\setlength{\tabcolsep}{4pt}
\begin{tabular}{@{}llrrrrrrr@{}}
\toprule
Model & Condition & SST-2 & MRPC & CoLA & RTE & NLI & MMLU & Avg.
\tabularnewline
\midrule
LLaDA & Unedited
& 97 & 65 & 78 & 79 & 80 & 59 & 76.3
\tabularnewline
LLaDA & MEMIT$^\dagger$, $N{=}2$
& 97 & 65 & 77 & 79 & 80 & 58 & 76.0
\tabularnewline
LLaDA & MEMIT$^\dagger$, $N{=}3$
& 97 & 65 & 79 & 79 & 81 & 59 & 76.7
\tabularnewline
LLaDA & MEMIT$^\dagger$, $N{=}4$
& 97 & 65 & 79 & 79 & 81 & 58 & 76.5
\tabularnewline
LLaDA & +Corr., $N{=}2$
& 97 & 65 & 78 & 79 & 81 & 58 & 76.3
\tabularnewline
LLaDA & +Corr., $N{=}3$
& 97 & 65 & 78 & 79 & 81 & 58 & 76.3
\tabularnewline
LLaDA & +Corr., $N{=}4$
& 97 & 65 & 78 & 79 & 81 & 58 & 76.3
\tabularnewline
\midrule
Dream & Unedited
& 89 & 53 & 71 & 74 & 69 & 67 & 70.5
\tabularnewline
Dream & MEMIT$^\dagger$, $N{=}2$
& 88 & 53 & 75 & 76 & 72 & 66 & 71.7
\tabularnewline
Dream & MEMIT$^\dagger$, $N{=}3$
& 90 & 52 & 76 & 76 & 65 & 67 & 71.0
\tabularnewline
Dream & MEMIT$^\dagger$, $N{=}4$
& 91 & 54 & 74 & 76 & 71 & 68 & 72.3
\tabularnewline
Dream & +Corr., $N{=}2$
& 86 & 56 & 78 & 75 & 68 & 68 & 71.8
\tabularnewline
Dream & +Corr., $N{=}3$
& 87 & 58 & 78 & 76 & 62 & 68 & 71.5
\tabularnewline
Dream & +Corr., $N{=}4$
& 91 & 53 & 75 & 76 & 66 & 68 & 71.5
\tabularnewline
\bottomrule
\end{tabular}
\caption{General-capability performance before and after editing.
$N$ denotes the target length of the edited facts. Scores are reported
as percentages, and Avg. is the unweighted average across the six
benchmarks. Identical evaluation examples are used across conditions.}
\label{tab:general-capability}
\end{table*}

Across all edited conditions, the six-task average remains within
0.5 percentage points of the unedited LLaDA and within 1.8 points of
the unedited Dream. Moreover, the averages of MEMIT$^\dagger$ and
+Corr. differ by at most 0.8 points, with no consistent direction.
Thus, neither editing nor the correction causes an observable loss of
general capability.

\subsection{Generation Quality and Accumulation Across Edits}
\label{app:locality-accumulation}

We evaluate generation quality using the standard \textsc{CounterFact}
fluency metric introduced with ROME \citep{meng2023locating}. This
metric is the weighted average of bi- and trigram entropies and decreases
when generation degenerates into repetition. For each condition, we
hold fixed 100 \textsc{CounterFact} generation prompts associated with
the edited subjects and score 64-token continuations. The same prompt
set is used before editing and at every edit batch size.

We then increase the number of simultaneous edits from 500 to 1,000
and 2,000. At each scale, we measure the six-task general-capability
average described above, generation fluency, and \textsc{CounterFact}
specificity. Table~\ref{tab:locality-scale} reports MEMIT$^\dagger$
and +Corr. under the same evaluation setup.

\begin{table*}[t]
\centering
\small
\setlength{\tabcolsep}{7pt}
\begin{tabular}{@{}llcccc@{}}
\toprule
Model & Metric & Pre-edit & 500 edits & 1,000 edits & 2,000 edits
\tabularnewline
\midrule
\multirow{3}{*}{LLaDA}
& General-capability avg.
& 76.3 & 76.2 / 76.5 & 76.8 / 77.2 & 76.8 / 76.8
\tabularnewline
& Fluency
& 5.37 & 5.38 / 5.42 & 5.46 / 5.45 & 5.49 / 5.38
\tabularnewline
& Specificity
& .177 & .192 / .183 & .185 / .188 & .174 / .176
\tabularnewline
\midrule
\multirow{3}{*}{Dream}
& General-capability avg.
& 70.5 & 70.5 / 68.0 & 69.5 / 68.5 & 72.0 / 70.8
\tabularnewline
& Fluency
& 4.43 & 4.68 / 4.65 & 4.45 / 4.66 & 4.64 / 4.68
\tabularnewline
& Specificity
& .174 & .201 / .257 & .277 / .240 & .279 / .276
\tabularnewline
\bottomrule
\end{tabular}
\caption{Model-preservation metrics as the number of edits increases.
At each edit count, paired entries report
MEMIT$^\dagger$/+Corr. General-capability avg. is the six-task average
from Table~\ref{tab:general-capability}; fluency is the weighted
bi- and trigram entropy; and specificity is measured on
\textsc{CounterFact}.}
\label{tab:locality-scale}
\end{table*}

For LLaDA, all three metrics remain close to their unedited values at
every scale. Dream likewise shows no systematic deterioration in
fluency or specificity. Its general-capability score under +Corr.
decreases at 500 and 1,000 edits but returns near the unedited value at
2,000 edits, rather than declining further with scale. Overall, we find no evidence of monotonic preservation degradation as
the edit batch grows to 2,000 facts. Relative to MEMIT$^\dagger$, the
correction remains close on most metrics, although Dream's
general-capability average is 1.0--2.5 points lower at the tested edit
scales.

\subsection{Ripple Effects}
\label{app:locality-ripple}

We further evaluate both the propagation of edited knowledge and the
preservation of related knowledge using
RippleEdits~\citep{cohen-etal-2024-evaluating}. Following its
original protocol, we evaluate 100 single edits from the Popular split,
whose high-frequency entities provide a challenging setting for ripple
effects. We report edit-verification success and accuracy on six
criteria: Logical Generalization (LG), Compositionality I and II
(CI and CII), Subject Aliasing (SA), Preservation (PV), and Relation
Specificity (RS). We use the same generation-based scoring pipeline for
the MDMs and the LLaMA-3 and Qwen2.5 ARM baselines.

Axis accuracies are computed only over edits that pass the benchmark's
edit-verification step. Because verification rates differ across
conditions, the axis scores are averaged over different subsets
($n=18$--$99$) and should not be compared without accounting for this
selection effect.

\begin{table*}[t]
\centering
\small
\setlength{\tabcolsep}{6pt}
\begin{tabular}{@{}lrrrrrrr@{}}
\toprule
Condition & Edit succ. & LG & CI & CII & SA & PV & RS
\tabularnewline
\midrule
LLaMA-3 MEMIT
& 44\% & .19 & .45 & .00 & 1.00 & 1.00 & .65
\tabularnewline
Qwen2.5 MEMIT
& 99\% & .06 & .39 & .00 & 1.00 & 1.00 & .41
\tabularnewline
\midrule
LLaDA MEMIT$^\dagger$
& 18\% & .33 & .67 & .00 & 1.00 & 1.00 & .65
\tabularnewline
LLaDA +Corr.
& 58\% & .14 & .83 & .00 & 1.00 & 1.00 & .73
\tabularnewline
Dream MEMIT$^\dagger$
& 20\% & .50 & .50 & .00 & 1.00 & 1.00 & .56
\tabularnewline
Dream +Corr.
& 70\% & .22 & .70 & .00 & 1.00 & 1.00 & .51
\tabularnewline
\bottomrule
\end{tabular}
\caption{\textsc{RippleEdits} results on 100 edits from the Popular
split. Edit succ. is the percentage of edits passing edit verification.
The six axis accuracies are averaged only over verified edits and
therefore use different subsets across conditions.}
\label{tab:ripple-effects}
\end{table*}

Among the edits verified under each condition, the MDM axis scores lie
within or above the range spanned by the two ARM baselines. In
particular, PV remains 1.00 in every condition, while MDM relation
specificity ranges from .51 to .73, compared with .41 to .65 for the
ARM baselines. These raw comparisons are descriptive rather than
controlled, however, because each condition is evaluated on a
different verified subset.

The benchmark also reproduces the multi-token editing difficulty.
Because entities in the Popular split are predominantly multi-token,
MEMIT$^\dagger$ often fails the initial edit-verification step.
The intermediate-state correction substantially increases verification
success on both MDMs.

LG appears lower after applying the correction, but this comparison is
confounded by the different verified subsets: +Corr. succeeds on many
more, and generally harder, edits. When evaluated only on facts edited
successfully by both conditions, LG is unchanged: both LLaDA conditions
score .33 and both Dream conditions score .50. These shared subsets are
small ($n=9$ for LLaDA and $n=8$ for Dream), because LG tests are
available for only a subset of the edited facts. Taken together, the correction substantially improves edit
verification on multi-token entities and does not reduce LG on the
small shared-success subsets. Because the remaining axis scores are
computed over different verified subsets, we do not draw a stronger
conclusion about its effects on related knowledge.

\section{Knowledge Editing versus Fine-Tuning}
\label{app:ft}

Fine-tuning (FT) provides a natural alternative to knowledge editing
for updating factual knowledge. However, standard FT distributes the
update across many parameters and may interfere with unrelated
knowledge. Constrained FT can reduce such interference, but prior
editing work has shown that it often overfits to the exact wording used
for training and generalizes poorly to paraphrases. We test whether the same
tradeoff arises in MDMs.

We compare constrained FT with MEMIT$^\dagger$ and +Corr. on LLaDA
using identical edit batches and evaluation prompts. To avoid
disadvantaging FT through a particular regularization choice, we sweep
its constraint strength and report its best-performing setting. At
this setting, unrelated general capability remains preserved: after
500 \textsc{CounterFact} edits, the six-task score from
Appendix~\ref{app:locality} is 76.2, compared with 76.3 for the
unedited model. The comparison therefore evaluates generalization
under a setting in which FT and editing both control unrelated
side effects.

\begin{table}[t]
\centering
\small
\setlength{\tabcolsep}{4pt}
\begin{tabular}{@{}llccc@{}}
\toprule
Dataset & Method & Efficacy & Gen. & Spec.
\tabularnewline
\midrule
\textsc{Kamel}, $N{=}2$
& FT & .60 & .21 & .18
\tabularnewline
& MEMIT$^\dagger$ & .60 & .36 & .12
\tabularnewline
& +Corr. & .87 & .52 & .12
\tabularnewline
\midrule
\textsc{CounterFact}
& FT & .99 & .28 & .22
\tabularnewline
& MEMIT$^\dagger$ & .91 & .56 & .19
\tabularnewline
& +Corr. & .95 & .64 & .18
\tabularnewline
\bottomrule
\end{tabular}
\caption{Comparison of constrained fine-tuning and knowledge editing
on LLaDA. \textsc{Kamel} uses 200 edits with two-token targets, and
\textsc{CounterFact} uses 500 edits. All methods are evaluated on the
same facts and prompts. Gen. and Spec. denote generalization and
specificity, respectively.}
\label{tab:editing-vs-ft}
\end{table}

Table~\ref{tab:editing-vs-ft} shows that FT can fit the original edit
prompt but generalizes substantially less well to alternative
expressions of the same fact. On both datasets, +Corr. achieves more
than twice the generalization of FT while maintaining comparable
specificity. The difference is especially clear on
\textsc{CounterFact}: FT reaches nearly perfect efficacy but retains
low generalization, indicating overfitting to the edit wording rather
than a failure to optimize the training examples.

These results reproduce the standard motivation for knowledge editing
in the MDM setting. When preservation of unrelated knowledge is held
approximately fixed, targeted editing transfers factual updates across
paraphrases substantially more effectively than constrained FT.

\section{Computational Resources}
All experiments were run on a single NVIDIA H200 GPU (144GB). The models in our main analyses have 7--8B parameters, while the
additional generality experiments span 355M--8B
(\S\ref{sec:setup}; Appendix~\ref{app:generality_models}). For the configurations used in our experiments, editing time is
approximately linear in the number of edited facts. LLaDA requires
about 8.7 seconds per fact (approximately 29 minutes for 200 facts),
while Dream requires about 4.5 seconds per fact (approximately
15 minutes for 200 facts). Editing 500 facts therefore takes
approximately 70 minutes for LLaDA and 40 minutes for Dream on a
single GPU.

\end{document}